\documentclass[10pt,twocolumn,letterpaper]{article}

\usepackage{cvpr}

\usepackage{graphicx}
\usepackage{amsmath}
\usepackage{amssymb}
\usepackage{booktabs}
\usepackage{multirow}
\usepackage{float}

\usepackage[pagebackref,breaklinks,colorlinks]{hyperref}

\usepackage[capitalize]{cleveref}
\crefname{section}{Sec.}{Secs.}
\Crefname{section}{Section}{Sections}
\Crefname{table}{Table}{Tables}
\crefname{table}{Tab.}{Tabs.}

\newcommand{\Mat}{\boldsymbol}
\newcommand{\Set}{\mathcal}

\newcommand{\real}{\mathbb{R}}

\def\cvprPaperID{4152} 
\def\confName{CVPR}
\def\confYear{2023}

\begin{document}

\title{U2NeRF: Unifying Unsupervised Underwater Image
Restoration and Neural Radiance Fields}

\author{
\hspace{-1mm}Vinayak Gupta\textsuperscript{1}$^{*}$, Manoj S\textsuperscript{1}$^{*}$, Mukund Varma T\textsuperscript{1}$^{\dagger}$\thanks{Equal contribution.}\,\,\,, Kaushik Mitra\textsuperscript{1}\thanks{Correspondence to: Mukund Varma T, and Kaushik Mitra.} \\ 
$^{1}$Indian Institute of Technology Madras \\
\texttt{\{vinayakguptapokal, mukundvarmat, manoj.s.2908\}@gmail.com},\\ \texttt{kmitra@ee.iitm.ac.in}
}

\maketitle

\begin{abstract}
    Underwater images suffer from colour shifts, low contrast, and haziness due to light absorption, refraction, scattering and restoring these images has warranted much attention. In this work, we present Unsupervised Underwater Neural Radiance Field (\textbf{U2NeRF}), a transformer-based architecture that learns to render and restore novel views conditioned on multi-view geometry simultaneously. Due to the absence of supervision, we attempt to implicitly bake restoring capabilities onto the NeRF pipeline and disentangle the predicted color into several components - scene radiance, direct transmission map, backscatter transmission map, and global background light, and when combined reconstruct the underwater image in a self-supervised manner. In addition, we release an Underwater View Synthesis (\textbf{UVS}) dataset consisting of 12 underwater scenes, containing both synthetically-generated and real-world data. Our experiments demonstrate that when optimized on a single scene, U2NeRF outperforms several baselines by as much LPIPS $\downarrow$11\%, UIQM $\uparrow$5\%, UCIQE $\uparrow$4\% (on average) and showcases improved rendering and restoration capabilities. Code will be made available upon acceptance. 
\end{abstract}

\section{Introduction}
\label{sec:intro}

Underwater images suffer from degradation due to poor, complex lighting conditions in water - more specifically due to light scattering, absorption and refraction~\cite{fu2022uncertainty}. Therefore, it is important to develop methods that can enhance underwater images, so that they are more suitable for visualization and downstream tasks like detection, tracking, etc. Recent advancements in deep learning has enabled great performance in several computer vision tasks~\cite{dosovitskiy2020vit, he2016deep}, including underwater image enhancement~\cite{chen2022domain, Fu_2022}. Most of these methods~\cite{desai2022aquagan, yin2021fmsnet} rely on synthetic training data due to the absence of large-scale real-world underwater image datasets with corresponding ground truth restored images. However, the synthesized data may not capture complex real-world degradation and thus suffer from domain shifts~\cite{chen2022domain}. More recently, ``zero-shot'' methods~\cite{kar2021zero, chai2022unsupervised} train a small image-specific network during test time and do not use any supervision other than the input image itself. However, they are not suitable for real-world applications due to a large number of optimization iterations at test time. 

\begin{figure}
\centering
\begin{minipage}[b]{0.28\textwidth}
  \centering
  \subfloat[Restoration]{\includegraphics[width=\textwidth]{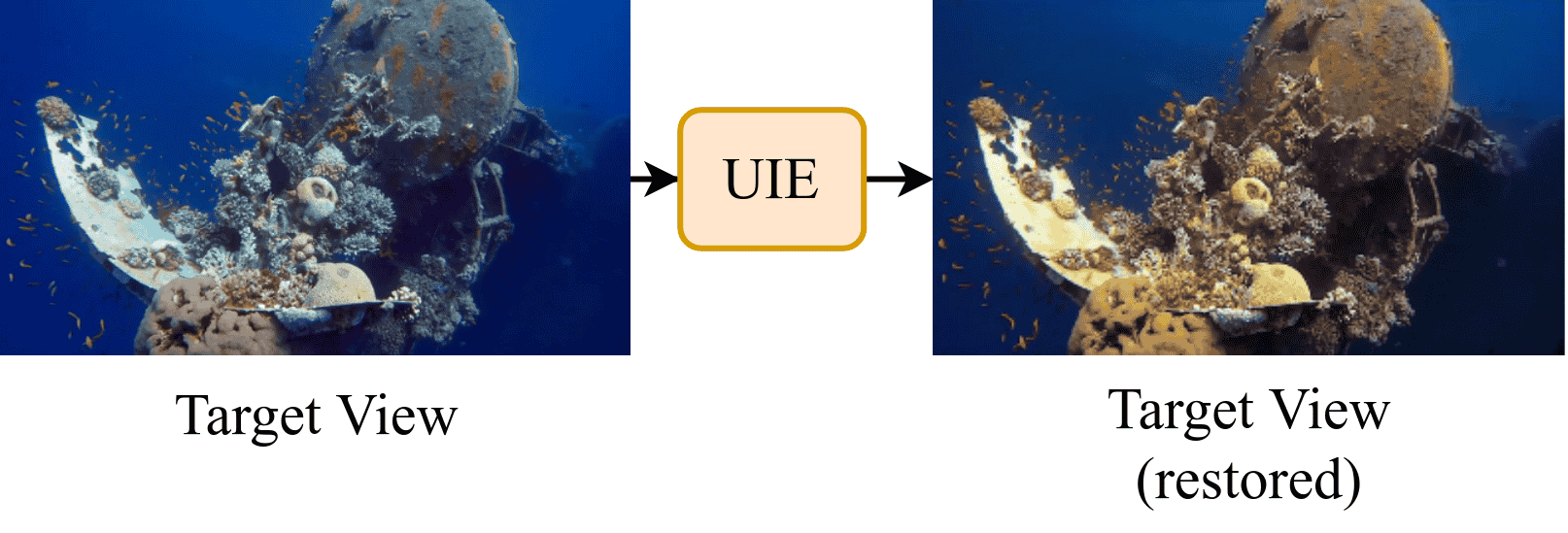}}
  \vfill
  \subfloat[Rendering]{\includegraphics[width=\textwidth]{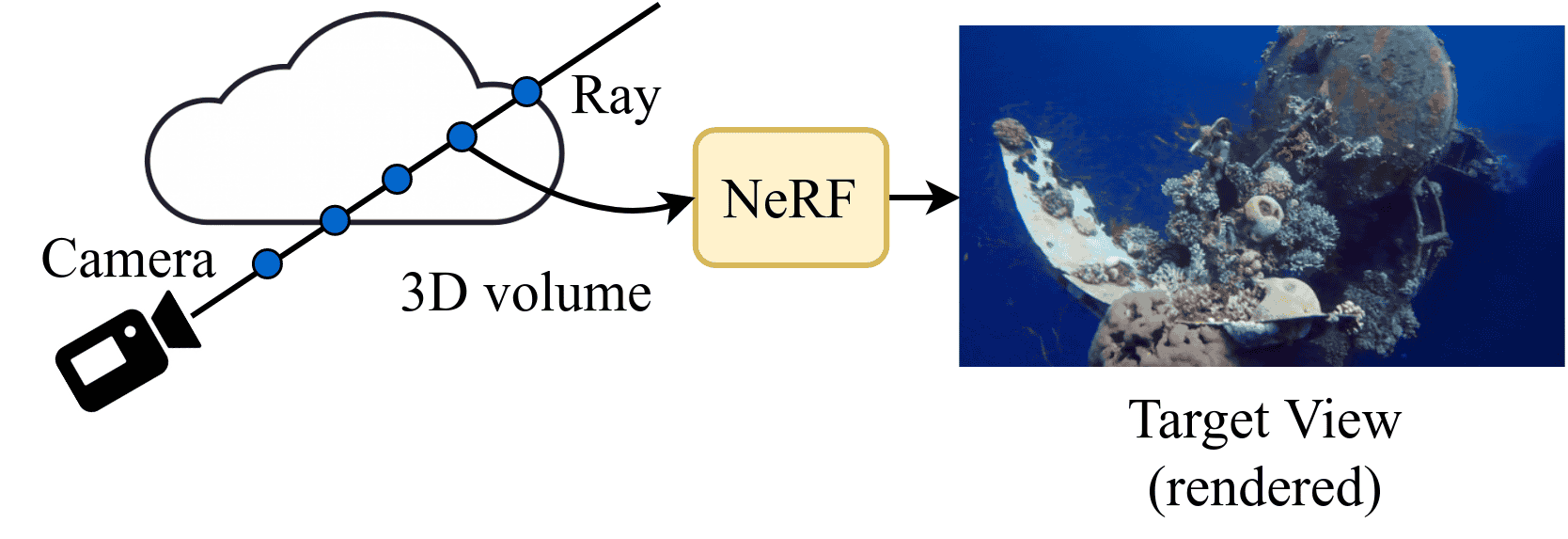}}
\end{minipage}
\subfloat[Ours]{\includegraphics[width=0.18\textwidth]{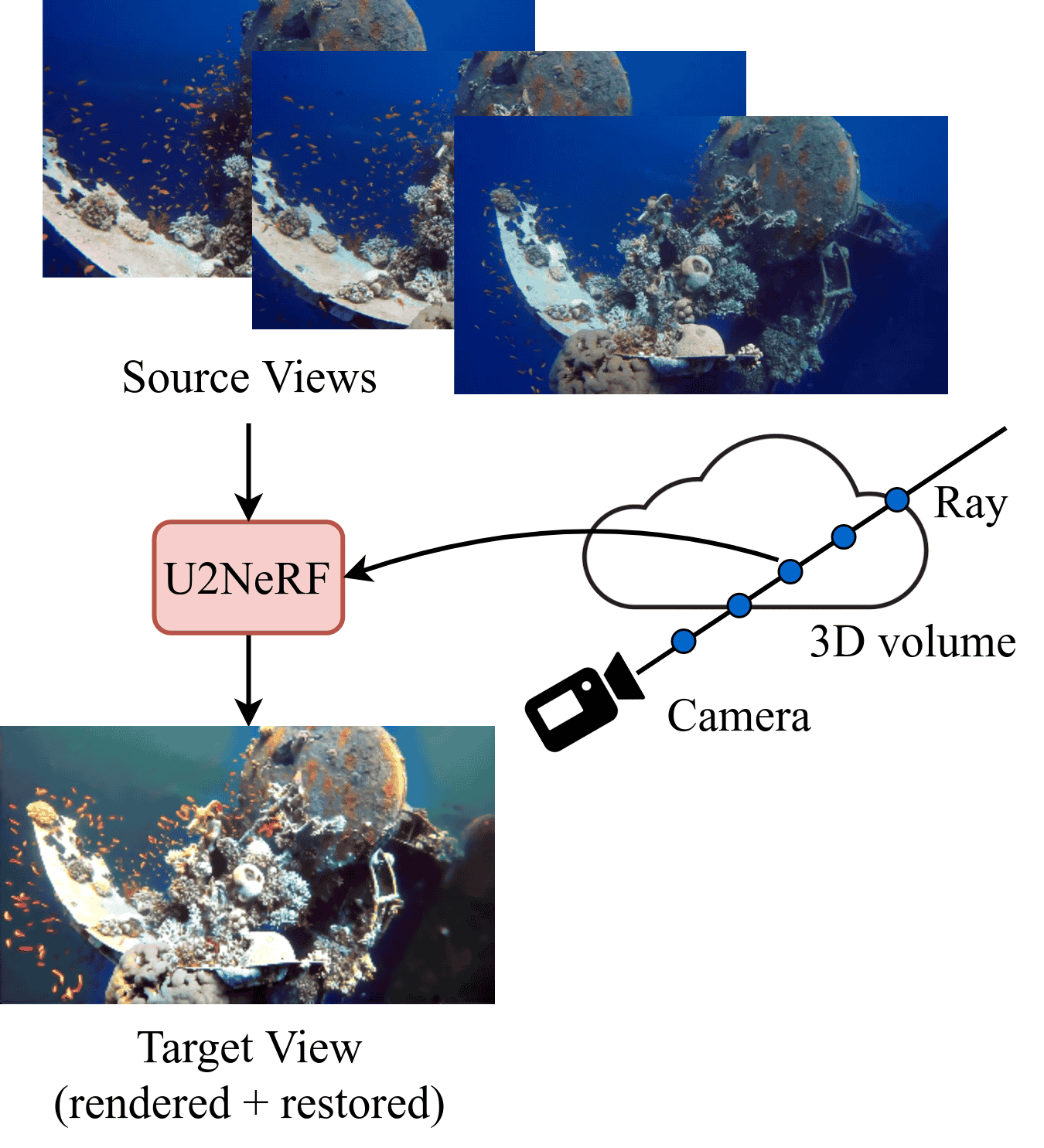}}
\caption{Unlike standalone methods like Radiance Fields (for rendering) and Image Enhancement (for restoration), our method U2NeRF simultaneously restores and renders an underwater scene. 
}
\end{figure}


Neural Radiance Fields~\cite{mildenhall2020nerf} and its follow-up works~\cite{barron2021mip, niemeyer2022regnerf, chen2022aug} have achieved remarkable success on novel view synthesis, generating photo-realistic, high-resolution, and view-consistent scenes. However, all these methods are trained on scenes containing clean, high-resolution images. Since NeRFs integrate information from multiple views, we hypothesize that these methods have a strong potential to be leveraged in multi-frame image restoration tasks - and in the context of this paper, underwater image enhancement. 

In this paper, we attempt to use NeRFs for simultaneous novel view rendering and restoration. However, most methods operate at a pixel level, limiting its capacity to automatically restore the predicted color. We demonstrate that by predicting image patches (rather than pixels), we provide sufficient spatial context for restoration. Motivated by ~\cite{chai2022unsupervised}, our method disentangles the predicted image patch into 4 components, namely scene radiance, direct transmission map, back scatter transmission map, and global background light. These components are later combined to reconstruct the original image, and along with suitable regularization enables our network to be trained in a self-supervised manner in the absence of clean ground truth image. Towards this end, we adapt the recently proposed Generalizable NeRF Transformer (GNT)~\cite{varma2022gnt}, which is composed of a \emph{view transformer} to aggregate multi-view information and render a novel view by composing colors along a ray using \emph{ray transformer}. Our method dubbed \textbf{U2NeRF} (Unsupervised Underwater Neural Radiance Field), trained in a fully unsupervised setting, learns to simultaneously render and restore a novel view. Our primary contributions can be summarized as follows:

\begin{enumerate}
    \item We extend the idea of radiance fields for the novel task of simultaneously rendering and restoring novel views, more specifically for underwater scenes. 
    \item Our proposed method U2NeRF, augments existing radiance fields with spatial awareness, and when combined with a physics-informed image formation model can successfully restore underwater images. 
    \item We contribute novel UVS Dataset consisting of 12 underwater scenes, containing both synthetically-generated and real-world data for novel view synthesis. Our proposed approach achieves best performance across perceptual (LPIPS $\downarrow$11\%) and color restoration metrics (UIQM $\uparrow$5\%, UCIQE $\uparrow$4\%). 
    \item Our results indicate that U2NeRF implicitly learns to generate physically meaningful image components, bringing us one step closer to using transformers as a universal modeling tool for graphics. 
\end{enumerate}

\section{Related Work}

\paragraph{Neural Radiance Fields. } NeRF introduced by \cite{mildenhall2020nerf} synthesizes consistent and photorealistic novel views by fitting each scene as a continuous 5D radiance field parameterized by an MLP. Since then, several works have improved NeRFs further. For example, 
Mip-NeRF~\cite{barron2021mip, barron2022mip} efficiently addresses scale of objects in unbounded scenes, Nex~\cite{Wizadwongsa2021NeX} models large view dependent effects, 
others~\cite{oechsle2021unisurf, yariv2021volume, wang2021neus} improve the surface representation, extend to dynamic scenes~\cite{park2021nerfies, park2021hypernerf, pumarola2021d} , introduce lighting and reflection modeling~ \cite{chen2021nerv, verbin2021ref}, or leverage depth to regress from few views~\cite{xu2022sinnerf, deng2022depth}. A recent work~\cite{pearl2022noiseaware} demonstrates the ability of NeRFs for burst denoising. Unlike other methods, our work aims to simultaneously render and restore a novel view, more specifically in the context of underwater scenes. 

\paragraph{Underwater Image Enhancement. }To compare our contribution to existing works for underwater image enhancement, two key factors must be considered: whether supervision is involved and whether the model refers to a certain physics model.

\begin{figure*}
\centering
\includegraphics[width=\textwidth]{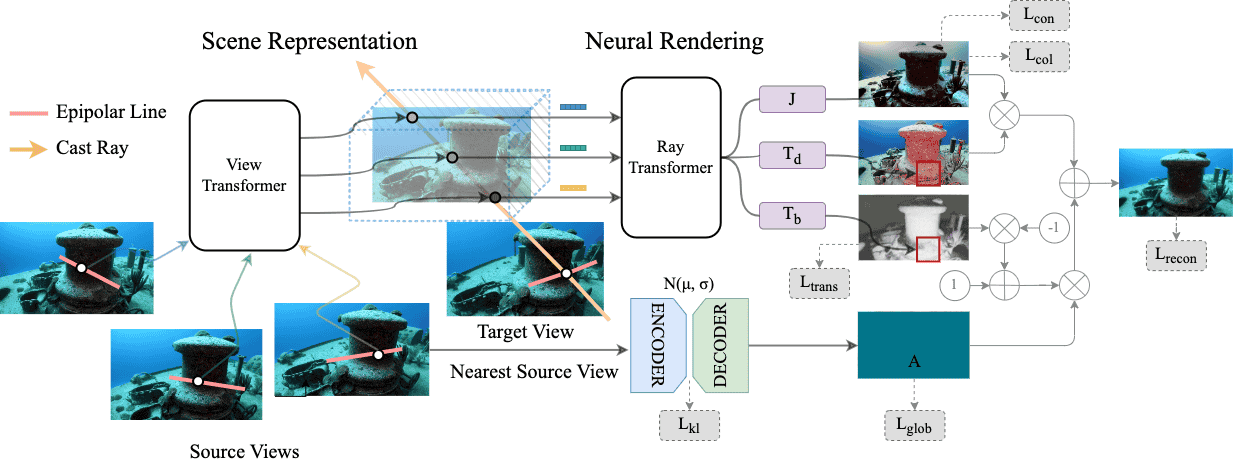}
\caption{Overview of U2NeRF: 1) Identify source views for a given target view, 2) Extract features for epipolar points using a trainable U-Net like model, 3) For each ray in the target view, sample points and directly predict a target patch disentangled into scene radiance, direct and backscatter transmission maps, and global background light. 4) The individual components are combined based on the image formation model to reconstruct the underwater image which is used as a self-supervision loss. }
\end{figure*}

\section{Unsupervised Underwater Neural Radiance Fields}
Our method U2NeRF extends GNT for the task for rendering and restoring novel views in underwater scenes. In this section, we first describe the preliminary of radiance fields, GNT, followed by a detailed description of our proposed method. 

\subsection{Preliminary}
\paragraph{Neural Radiance Fields. } NeRFs~\cite{mildenhall2020nerf} converts multi-view images into a radiance field and interpolates novel views by re-rendering the radiance field from a new angle. Technically, NeRF models the underlying 3D scene as a continuous radiance field $\mathcal{F}: (\Mat{x}, \Mat{\theta}) \mapsto (\Mat{c}, \sigma)$ parameterized by a Multi-Layer Perceptron (MLP) $\Mat{\Theta}$, which maps a spatial coordinate $\Mat{x} \in \real^{3}$ together with the viewing direction $\Mat{\theta} \in [-\pi, \pi]^{2}$ to a color $\Mat{c} \in \real^{3}$ plus density $\sigma \in \real_{+}$ tuple.
To form an image, NeRF performs the ray-based rendering, where it casts a ray $\Mat{r} = (\Mat{o}, \Mat{d})$ from the optical center $\Mat{o} \in \real^{3}$ through each pixel (towards direction $\Mat{d} \in \real^3$), and then leverages volume rendering~\cite{kajita1984ray} to compose the color and density along the ray between the near-far planes:
\begin{align}
\label{eqn:rend}
    \Mat{C}(\Mat{r} \vert \Mat{\Theta}) = \int_{t_{n}}^{t_f} T(t)\sigma(\Mat{r}(t))\Mat{c}(\Mat{r}(t), \Mat{d}) dt, \\ \text{ where } T(t) = \exp\left({-\int_{t_{n}}^{t} \sigma(s) ds}\right) \notag
\end{align}
where $\Mat{r}(t) = \Mat{o} + t \Mat{d}$, $t_n$ and $t_f$ are the near and far planes respectively. In practice, the Eqn. \ref{eqn:rend} is numerically estimated using quadrature rules~\cite{max1995optical}. Given images captured from surrounding views with known camera parameters, NeRF fits the radiance field by maximizing the likelihood of simulated results.
Suppose we collect all pairs of rays and pixel colors as the training set $\Set{D} = \{(\Mat{r}_{i}, \widehat{\Mat{C}}_{i})\}_{i=1}^{N}$, where $N$ is the total number of rays sampled, and $\widehat{\Mat{C}}_{i}$ denotes the ground-truth color of the $i$-th ray, then we train the implicit representation $\Mat{\Theta}$ via the following loss function:
\begin{equation} \label{eqn:nerf_loss}
    \mathcal{L}(\Mat{\Theta} \vert \Set{R}) = \mathbb{E}_{(\Mat{r}, \widehat{\Mat{C}}) \in \Set{D}} \lVert \Mat{C}(\Mat{r} \lvert \Mat{\Theta}) - \widehat{\Mat{C}}(\Mat{r}) \rVert_2^{2},
\end{equation}

\paragraph{Generalizable NeRF Transformer. } 
GNT \cite{varma2022gnt} considers the problem of novel view synthesis as a two stage information process: the multi-view image feature fusion, followed by the sampling-based rendering integration. It is composed of (i) \textit{view transformer} to aggregate pixel-aligned image features from corresponding epipolar lines to predict coordinate-wise features, (ii) \textit{ray transformer} to compose coordinate-wise point features along a traced ray via attention mechanism. More formally, the entire operation can be summarized as follows:

\begin{footnotesize}
\begin{align}
    \mathcal{F}(\Mat{x}, \Mat{\theta}) = \operatorname{View-Transformer}(\Mat{F}_1(\Pi_1(\Mat{x}), \Mat{\theta}), \cdots, \\ \Mat{F}_N(\Pi_N(\Mat{x}), \Mat{\theta})) \notag,
\end{align}
\end{footnotesize}

where $\operatorname{View-Transformer}(\cdot)$ is a transformer encoder,  $\Pi_i(\Mat{x})$ projects position $\Mat{x} \in \real^3$ onto the $i$-th image plane by applying extrinsic matrix, and $\Mat{F}_i(\Mat{z}, \Mat{\theta}) \in \real^d$ computes the feature vector at position $\Mat{z} \in \real^2$ via bilinear interpolation on the feature grids. The multi-view aggregated point features are fed into the ray transformer, and the output from the ray transformer is fed into a view transformer and this process is repeated where the view transformer and ray transformer are stacked alternatively. Then the features from the last ray transformer are pooled to extract a single ray feature to predict the target pixel color. 

\begin{footnotesize}
\begin{align}
    \Mat{C}(\Mat{r}) = \operatorname{MLP} \circ \operatorname{Mean} \circ \operatorname{Ray-Transformer}(\mathcal{F}(\Mat{o} + t_1 \Mat{d}, \Mat{\theta}), \cdots, \\ \mathcal{F}(\Mat{o} + t_M \Mat{d}, \Mat{\theta})) \notag,
\end{align}
\end{footnotesize}

where $t_1, \cdots, t_M$ are uniformly sampled  between near and far plane, $\operatorname{Ray-Transformer}$ is a standard transformer encoder.


\subsection{Baking Restoration Capabilities onto U2NeRF}

NeRF represents 3D scene as a radiance field $\mathcal{F}: (\Mat{x}, \Mat{\theta}) \mapsto (\Mat{c}, \sigma)$, where each spatial coordinate $\Mat{x} \in \real^{3}$ together with the viewing direction $\Mat{\theta} \in [-\pi, \pi]^{2}$ is mapped to a color $\Mat{c} \in \real^{3}$ plus density $\sigma \in \real_{+}$ tuple. However, a single pixel does not provide sufficient context for automatic restoration. In our work, we first adapt GNT to render an image patch of size $p$. The final ray feature obtained from the ray transformer block is passed on to a sequence of convolution and upsampling layers. Motivated by ~\cite{chai2022unsupervised}, we disentangle the underwater image into several components - scene radiance ($J$), global background light ($A$) and degradation components - direct and back scatter transmission maps ($T_{D}$, $T_{B}$) that account for attenuation and light reflection respectively. These individual components can be combined to reconstruct the original image $I$ at pixel $i$ as: 
\begin{align}
    \label{eqn:physics}
    I(i) = J(i)T_{D} (i) + (1 - T_{B} (i))A
\end{align}
This enables our network to be trained in a fully self-supervised manner in the absence of ground truth image. To predict $J$, $T_{D}$, and $T_{B}$, we initialize separate output heads to project the final ray feature to the desired patch size. Since $A$ is independent of the input image content, we pass the nearest source image from the target view direction onto a Variational AutoEncoder (VAE) to estimate global background light. In addition to the photometric loss given in Eqn. \ref{eqn:nerf_loss} ($\mathcal{L}_{\text{rec}}$), we (1) minimize the difference between encoded feature $z$ and latent code sampled from Gaussian $\hat{z}$ in the vae ($\mathcal{L}_{\text{kl}}$), (2) minimize the difference between the saturation and brightness of the predicted scene radiance to reduce haze ($\mathcal{L}_{\text{con}}$), (3) minimize the potential color deviations in the scene radiance($\mathcal{L}_{\text{col}}$), (4) ensure constant back-scatter coefficients  ($\mathcal{L}_{\text{trans}}$) across channels, (5) enforce constant global background light by minimizing variance within each local neighbourhood ($\mathcal{L}_{\text{glob}}$) as proposed in the original paper~\cite{chai2022unsupervised}. Together, the network is trained to optimize:
\begin{align}
    \label{eqn:loss}
    \mathcal{L} = \lambda_1\mathcal{L}_{rec}, + \lambda_2\mathcal{L}_{con} + \lambda_3\mathcal{L}_{col} + \lambda_4\mathcal{L}_{kl} \\ + \lambda_5\mathcal{L}_{trans} + \lambda_6\mathcal{L}_{glob} \notag
\end{align}
where $\lambda$ indicates the weight for each loss term.

\section{Experiments}
We conduct extensive experiments to compare U2NeRF with several baseline methods for
novel view synthesis and restoration in the context of underwater scenes. We first provide qualitative and quantitative results in the single scene training setting, followed by extending our approach for generalization to unseen scenes.

\subsection{Implementation Details}
\paragraph{Source and Target view sampling. } As described in~\cite{wang2021ibrnet}, we construct a training pair of source and target views by first selecting a target view, then identifying a pool of k × N nearby views, from which N views are randomly sampled as source views. This sampling strategy simulates various view densities during training and therefore helps the network generalize better. During training, the values for k and N are uniformly sampled at random from [1-3] and [8-12], respectively.
\vspace{-4mm}
\paragraph{Network Architecture. } To extract features from the source views, we use a U-Net like architecture with a ResNet34 encoder, followed by two up-sampling layers as decoder~\cite{wang2021ibrnet}. Each view transformer block contains a single-headed cross-attention layer while the ray transformer block contains a multi-headed self-attention layer with four heads. The outputs from these attention layers are passed onto corresponding feedforward blocks with a Rectified Linear Unit (RELU) activation and a hidden dimension of 256. A residual connection is applied between the pre-normalized inputs (LayerNorm) and outputs at each layer. For all our single scene experiments, we alternatively stack 4 view and ray transformer blocks while our larger generalization experiments use 8 blocks each. All transformer blocks (view and ray) are of dimension 64. We set the patch size $p$ as 4 for all our experiments to arrive at a balance between performance and network complexity. The VAE network contains 4 convolution layers with dimensions [16, 32, 64, 128] in the encoder, each followed by relu activation. The encoded input is projected to 100 dimension feature vector before Gaussian re-sampling. The sampled Gaussian latent is then passed onto a 3-layer decoder network with dimensions [128, 64, 32] to predict global background light $A$.

\begin{figure*}[!t]
\centering
\begin{subfigure}[t]{0.18\textwidth}
  \centering
  \includegraphics[width=1.0\linewidth]{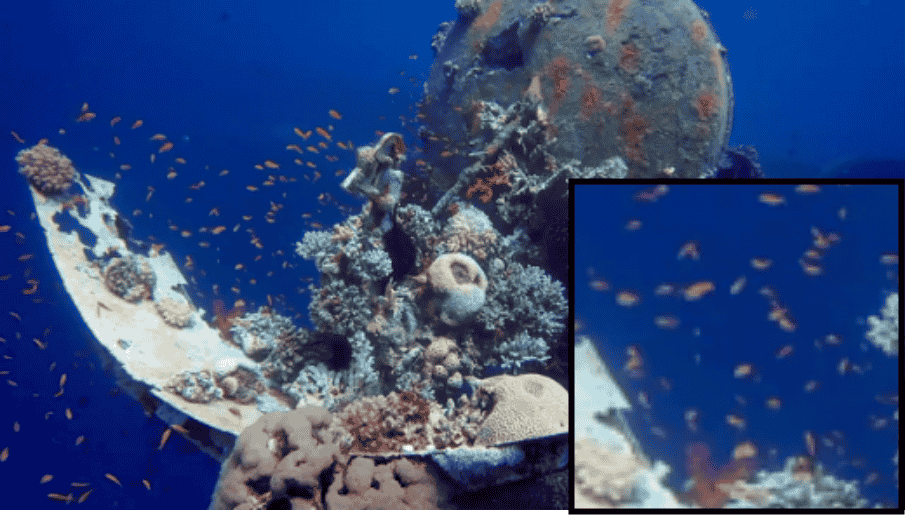}
\label{fig:debris_uw} 
\end{subfigure}%
\begin{subfigure}[t]{0.18\textwidth}
  \centering
  \includegraphics[width=1.0\linewidth]{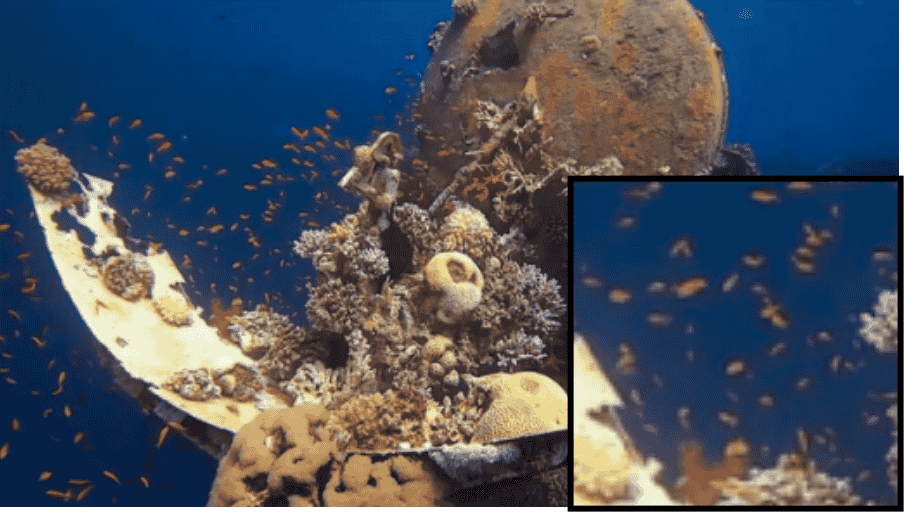}
\label{fig:debris_uies} 
\end{subfigure}%
\begin{subfigure}[t]{0.18\textwidth}
  \centering
  \includegraphics[width=1.05\linewidth]{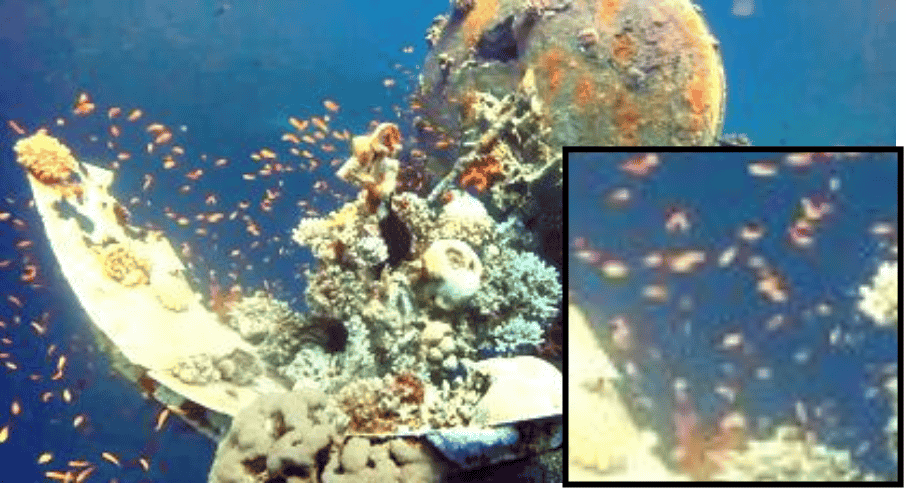}
\label{fig:debris_upifm} 
\end{subfigure}%
\begin{subfigure}[t]{0.18\textwidth}
  \centering
  \includegraphics[width=1.0\linewidth]{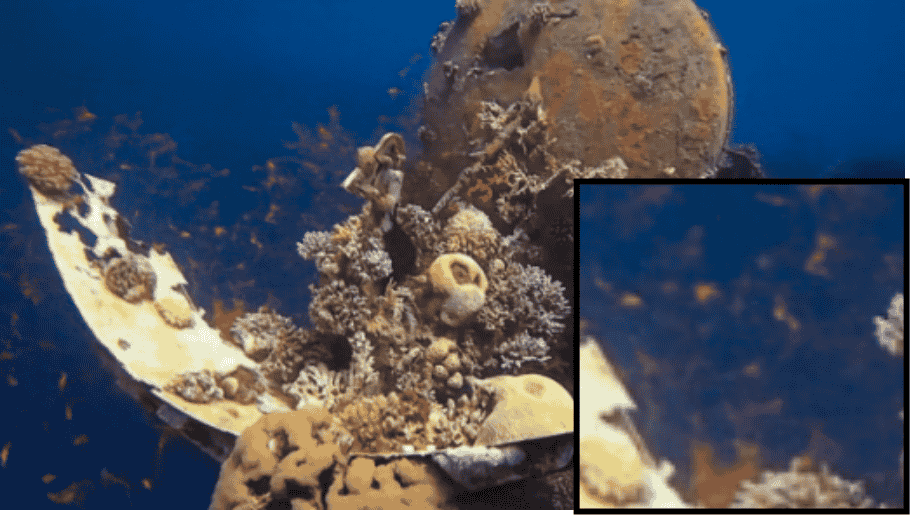}
\label{fig:debris_nerf_clean} 
\end{subfigure}%
\begin{subfigure}[t]{0.18\textwidth}
  \centering
  \includegraphics[width=1.0\linewidth]{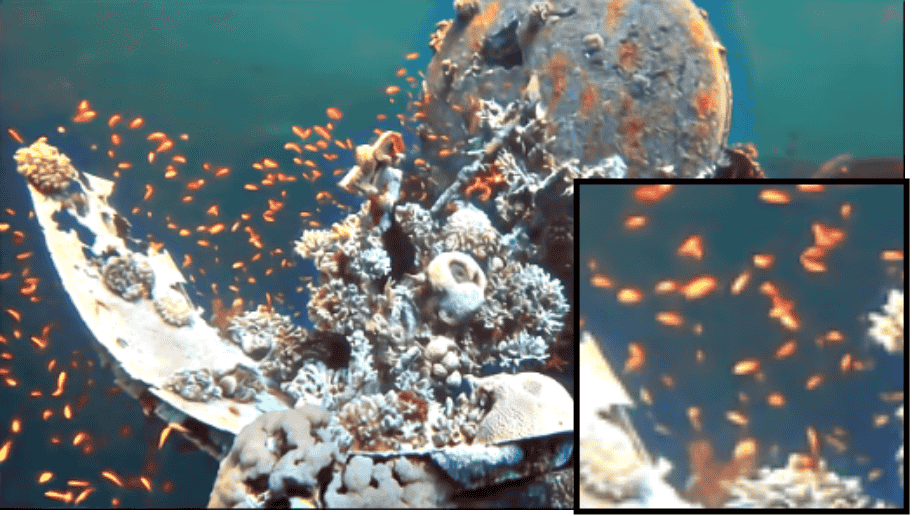}
\label{fig:debris_u2nerf} 
\end{subfigure}%
\setcounter{subfigure}{0}
\centering
\begin{subfigure}[t]{0.18\textwidth}
  \centering
  \includegraphics[width=1.0\linewidth]{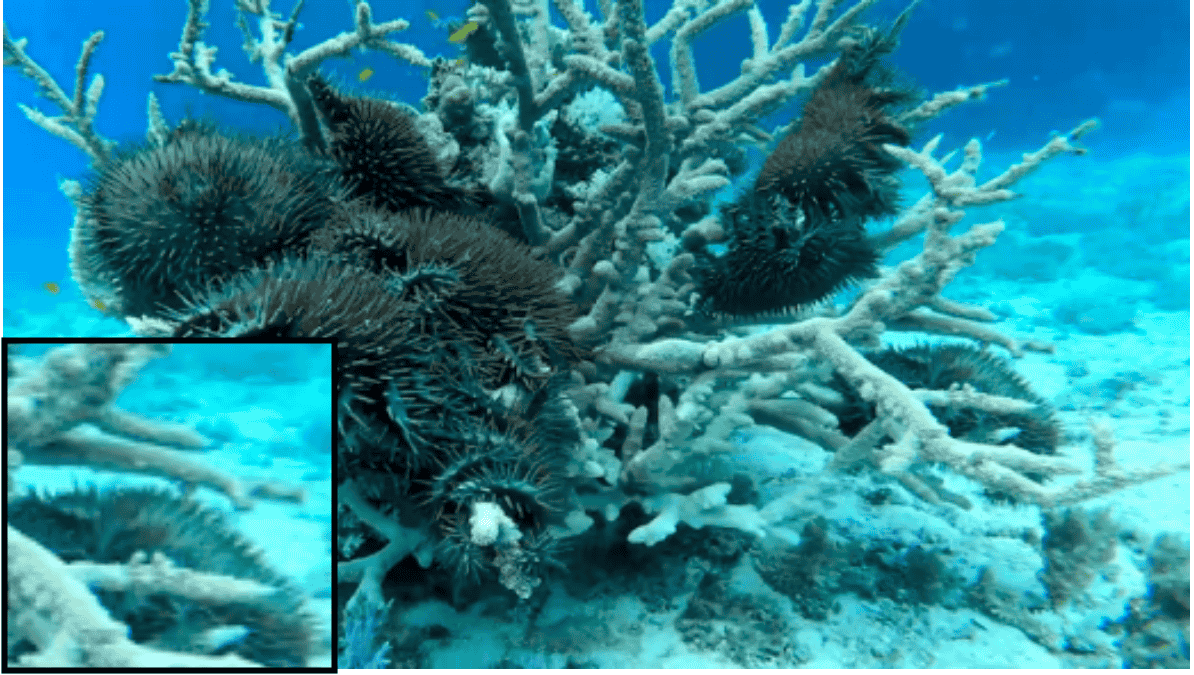}
\label{fig:starfish_uw} 
\end{subfigure}%
\begin{subfigure}[t]{0.18\textwidth}
  \centering
  \includegraphics[width=1.0\linewidth, height=17.8mm]{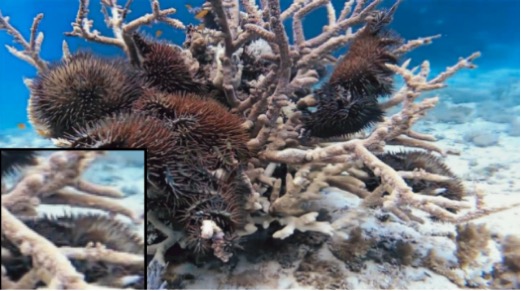}
\label{fig:starfish_uies} 
\end{subfigure}%
\begin{subfigure}[t]{0.18\textwidth}
  \centering
  \includegraphics[width=1.0\linewidth, height=17.8mm]{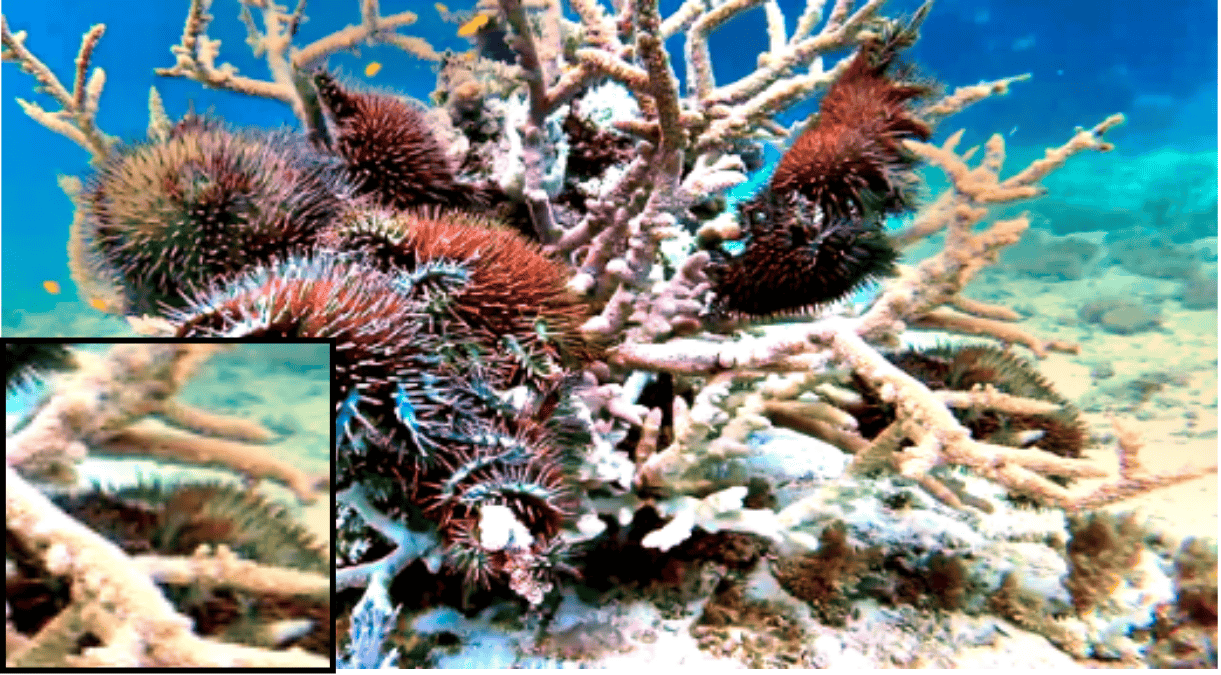}
\label{fig:starfish_upifm} 
\end{subfigure}%
\begin{subfigure}[t]{0.18\textwidth}
  \centering
  \includegraphics[width=1.0\linewidth]{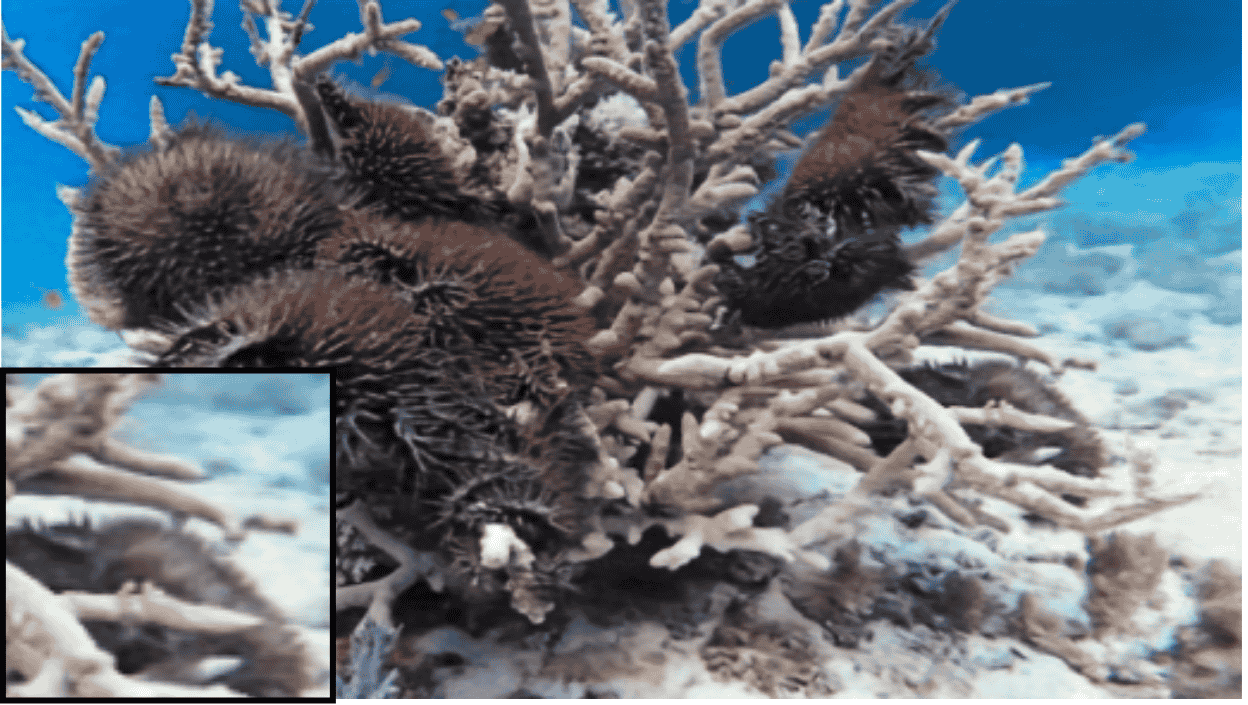}
\label{fig:starfish_nerf_clean} 
\end{subfigure}%
\begin{subfigure}[t]{0.18\textwidth}
  \centering
  \includegraphics[width=1.0\linewidth]{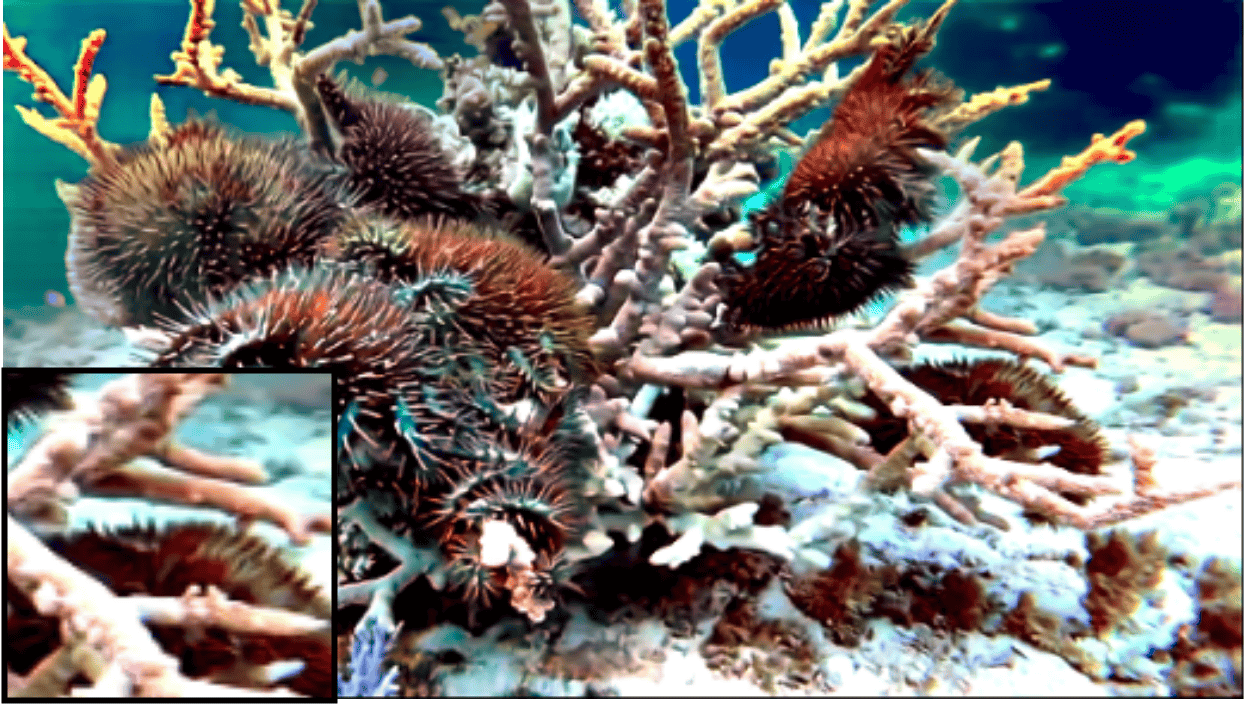}
\label{fig:starfish_u2nerf} 
\end{subfigure}%
\setcounter{subfigure}{0}
\begin{subfigure}[t]{0.18\textwidth}
  \centering
  \includegraphics[width=1.0\linewidth]{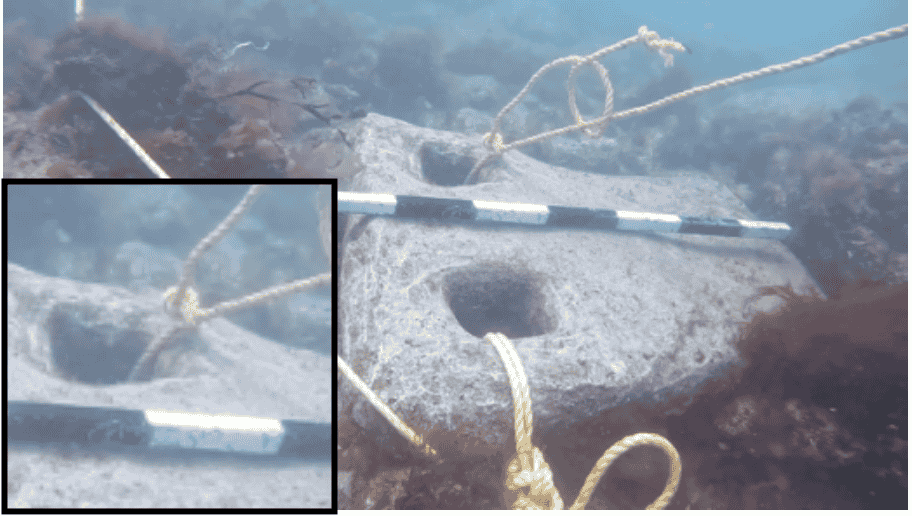}
\caption{Underwater Image}
\label{fig:dwarka26_uw} 
\end{subfigure}%
\begin{subfigure}[t]{0.18\textwidth}
  \centering
  \includegraphics[width=1.0\linewidth]{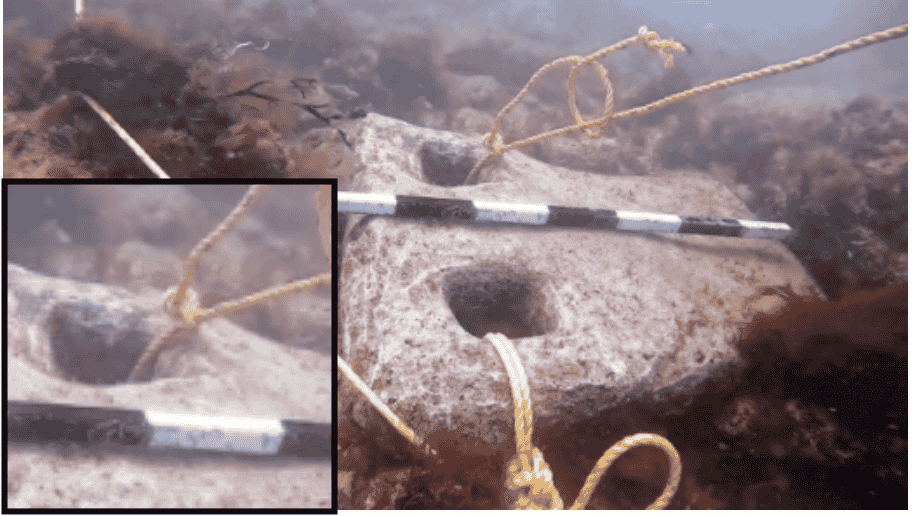}
\caption{UIESS}
\label{fig:dwarka26_uies} 
\end{subfigure}%
\begin{subfigure}[t]{0.18\textwidth}
  \centering
  \includegraphics[width=1.0\linewidth, , height=17.8mm]{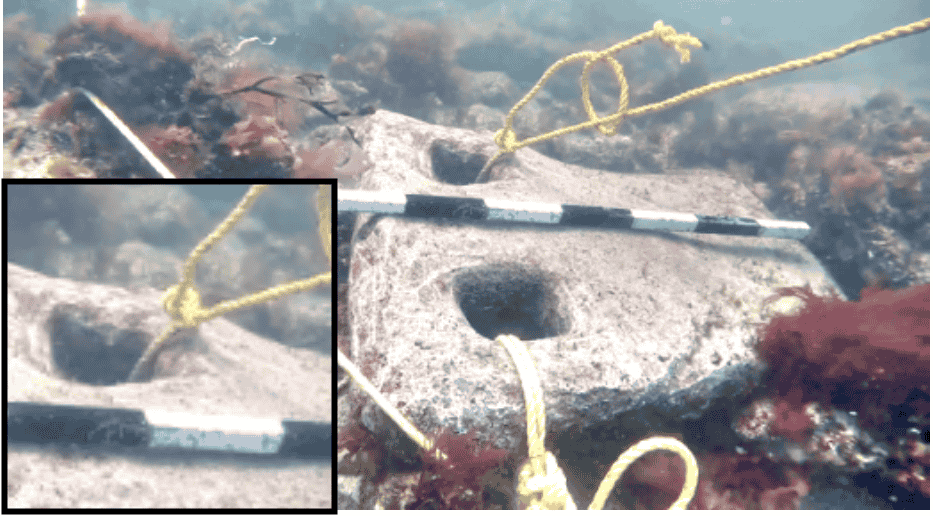}
\caption{UPIFM}
\label{fig:dwarka26_upifm} 
\end{subfigure}%
\begin{subfigure}[t]{0.18\textwidth}
  \centering
  \includegraphics[width=1.0\linewidth]{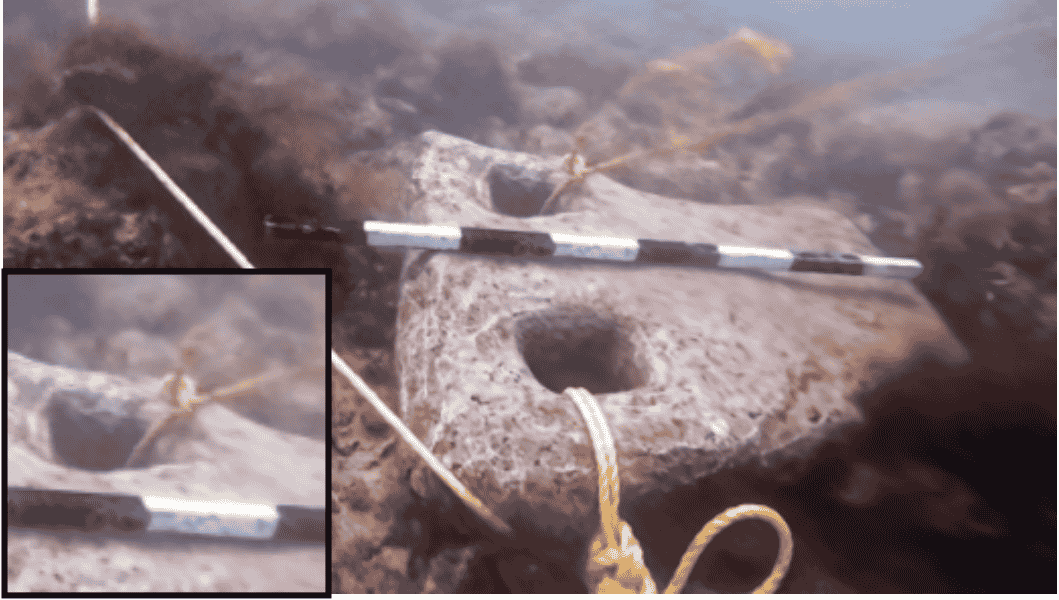}
\caption{NeRF + Clean}
\label{fig:dwarka26_nerf_clean} 
\end{subfigure}%
\begin{subfigure}[t]{0.18\textwidth}
  \centering
  \includegraphics[width=1.0\linewidth]{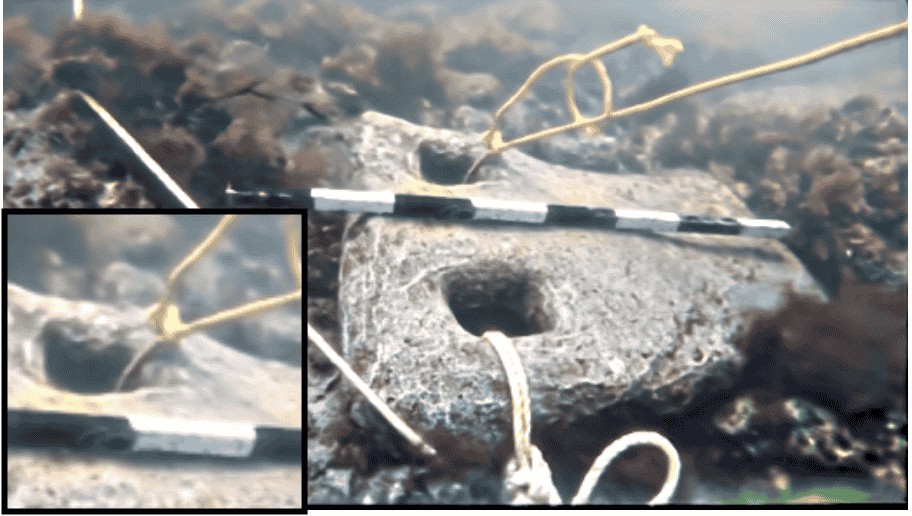}
\caption{U2NeRF}
\label{fig:dwarka26_u2nerf} 
\end{subfigure}%
\caption{Qualitative results for single-scene rendering. In the Debris scene (row-1), U2NeRF is able to successfully recover and restore fishes, and enhance its visibility. In the Starfish scene (row-2), U2NeRF reconstructs edges with greater detail and even comparable to the non-rendering baselines. In scene2 from `hard' split (row-3), U2NeRF renders complex, moving structures (rope) with higher visual quality. }
\label{fig:single}
\end{figure*}

\paragraph{Training / Inference Details. } We train both the feature extraction network and U2NeRF end-to-end on datasets of multi-view posed images using the Adam optimizer to minimize the loss given in Eqn. \ref{eqn:loss}. We empirically set the values of $\lambda$ to $\lambda_1$ = 1, $\lambda_2$ = 0.1, $\lambda_3$ = 1, $\lambda_4$ = 1, $\lambda_5$ = 0.1, $\lambda_6$ = 0.1. The base learning rates for the feature extraction network and U2NeRF are $10^{-3}$ and $5 \times 10^{-4}$ respectively, which decay exponentially over training steps. During finetuning, we optimize the feature extraction network and U2NeRF using a smaller learning rate of $5 \times 10^{-4}$ and $2 \times 10^{-4}$. For the single scene and cross scene generalization experiments, we train U2NeRF for 250,000 steps with 512 rays sampled in each iteration while during fine-tuning on each scene, the pre-trained network is only fine-tuned for 50,000 steps with 256 rays sampled in each iteration. Unlike most NeRF methods, we do not use separate coarse, fine networks and therefore to bring GNT to a comparable experimental setup, we sample 192 coarse points per ray across all experiments (unless otherwise specified). 

\begin{table*}
  \centering
  \begin{subtable}[t]{0.33\textwidth}
  \centering
  \resizebox{\columnwidth}{!}{
  \begin{tabular}{lccc}
    \toprule
    Models & PSNR$\uparrow$ & SSIM$\uparrow$ & LPIPS$\downarrow$ \\
    \midrule
    UPIFM & 12.883 & 0.329 & 0.399 \\
    UIESS & 18.818 & 0.790 & 0.174\\
    \midrule
    NeRF & 12.283 & 0.558 & 0.360\\
    NeRF + Clean & 17.948 & 0.741 & 0.297\\
    \midrule
    U2NeRF & 13.978 & 0.594 & 0.230\\
    \bottomrule
  \end{tabular}%
  }
  \caption{Easy Split}
  \label{tab:single_scene_easy}
  \end{subtable}
  \centering
  \begin{subtable}[t]{0.33\textwidth}
  \centering
  \resizebox{\columnwidth}{!}{
  \begin{tabular}{lccc}
    \toprule
    Models & UIQM$\uparrow$ & UCIQE$\uparrow$ & LPIPS (gray)$\downarrow$ \\
    \midrule
    UPIFM & 1.424 & 32.940 & - \\
    UIESS & 1.136 & 30.534 & - \\
    \midrule
    NeRF & 0.501 & 31.622 & 0.208\\
    NeRF + Clean & 0.865 & 31.054 & 0.223\\
    Clean + NeRF & 0.858 & 30.336 & 0.198\\
    \midrule
    U2NeRF & 1.570 & 32.556 & 0.174\\
    \bottomrule
  \end{tabular}%
  }
  \caption{Medium Split}
  \label{tab:single_scene_medium}
  \end{subtable}
  \begin{subtable}[t]{0.33\textwidth}
  \centering
  \resizebox{\columnwidth}{!}{
  \begin{tabular}{lccc}
    \toprule
    Models & UIQM$\uparrow$ & UCIQE$\uparrow$ & LPIPS (gray)$\downarrow$ \\
    \midrule
    UPIFM & 1.182 & 28.537 & -\\
    UIESS & 0.649 & 27.161 & -\\
    \midrule
    NeRF & 0.463 & 18.370 & 0.334\\
    NeRF + Clean & 0.486 & 27.453 & 0.328\\
    Clean + NeRF & 0.456 & 26.530 & 0.292\\
    \midrule
    U2NeRF & 1.100 & 26.788 & 0.260\\
    \bottomrule
  \end{tabular}}
  \caption{Hard Split}
  \label{tab:single_scene_hard}
  \label{tab:single_scene}
  \end{subtable}
  \caption{Comparison of U2NeRF against baseline methods for single-scene rendering on the UVS dataset}
\end{table*}

\paragraph{Metrics. } To evaluate our method's rendering and restoration quality, we use widely adopted metrics: Peak Signal-to-Noise Ratio (PSNR), Structural Similarity Index Measure (SSIM)~\cite{wang2004image}, Learned Perceptual Image Patch Similarity (LPIPS)~\cite{zhang2018unreasonable}, Underwater Image Quality Measurement (UIQM)~\cite{panetta2015human}, and the Underwater Color Image Quality Evaluation Metric (UCIQE)~\cite{yang2015underwater}. We report the averages of each metric over different views in each scene and across multiple scenes in each dataset. Following previous works~\cite{mildenhall2020nerf, varma2022gnt}, we compute the PSNR, SSIM, LPIPS scores between the rendered and ground truth restored views in the case of synthetic scenes, and the UIQM, UCIQE scores to evaluate real underwater scenes, with no reference restored image. In the case of real world data, we additionally report LPIPS scores between the gray scale image of rendered and restored target views using \cite{chen2022domain} (to remove color differences) and quantitatively measure the rendering capabilities of different methods. 

\begin{figure*}[!t]
\centering
\rotatebox{90}{\footnotesize Underwater Image}
\begin{subfigure}[t]{0.22\textwidth}
  \centering
  \includegraphics[width=1.0\linewidth]{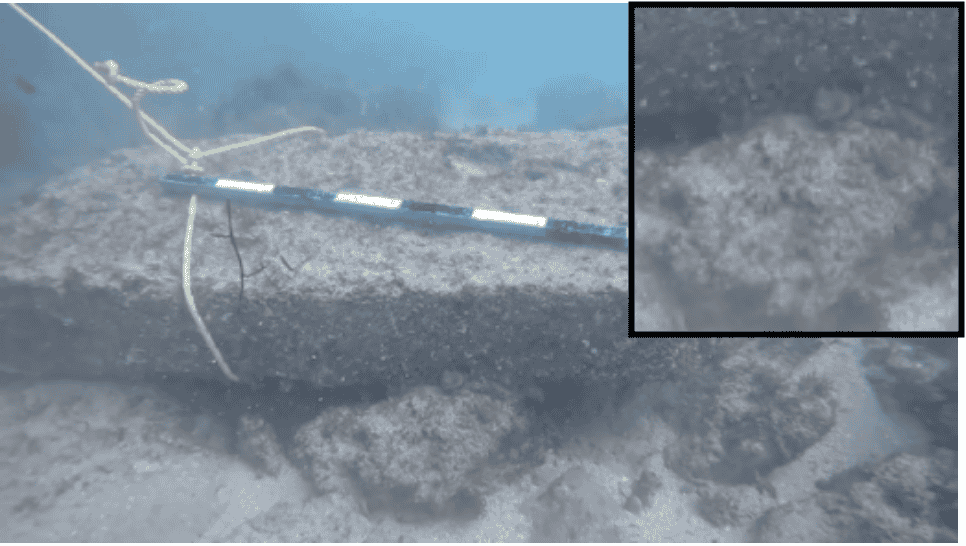}
\label{fig:debris_uw} 
\end{subfigure}%
\begin{subfigure}[t]{0.22\textwidth}
  \centering
  \includegraphics[width=1.0\linewidth]{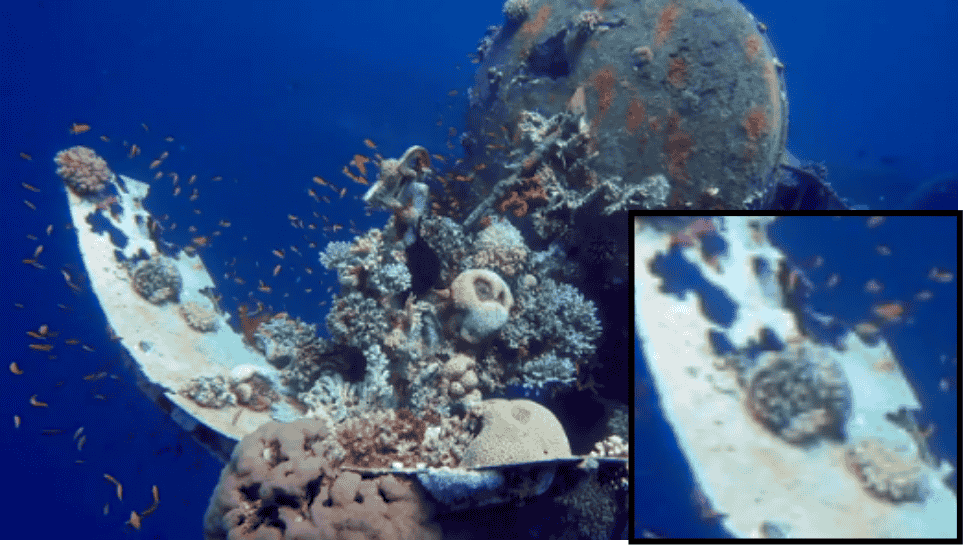}
\label{fig:debris_uies} 
\end{subfigure}%
\begin{subfigure}[t]{0.22\textwidth}
  \centering
  \includegraphics[width=1.0\linewidth]{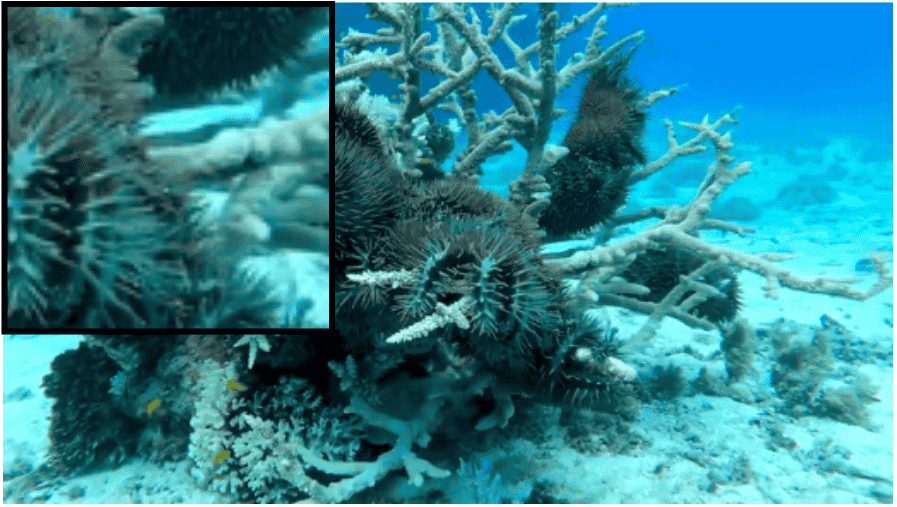}
\label{fig:debris_upifm} 
\end{subfigure}%
\begin{subfigure}[t]{0.22\textwidth}
  \centering
  \includegraphics[width=1.0\linewidth]{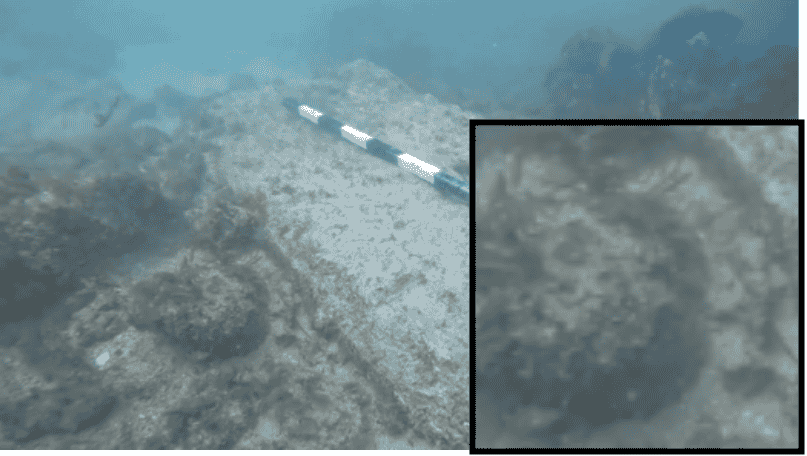}
\label{fig:debris_nerf_clean} 
\end{subfigure}%

\setcounter{subfigure}{0}
\centering
\rotatebox{90}{\footnotesize U2NeRF (Gen)}
\begin{subfigure}[t]{0.22\textwidth}
  \centering
  \includegraphics[width=1.0\linewidth]{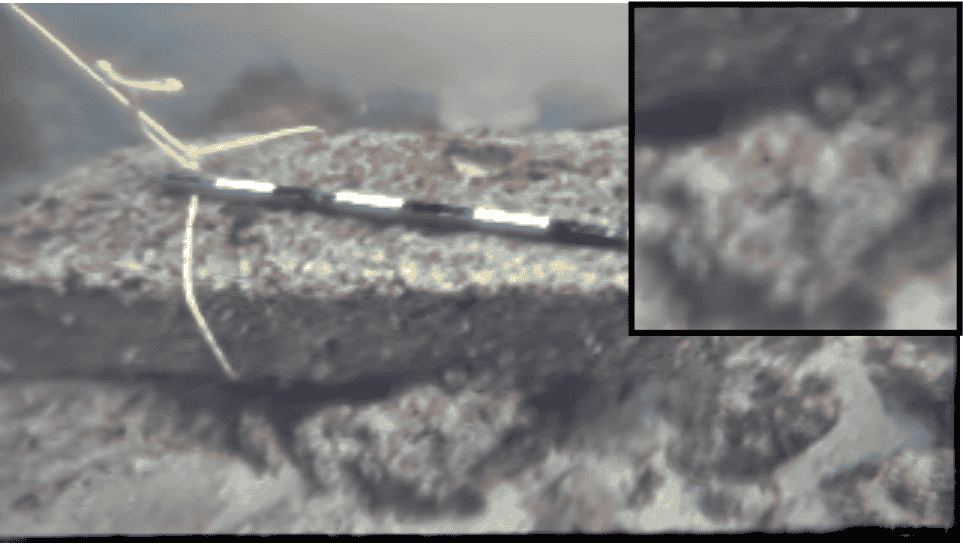}
\label{fig:starfish_uw} 
\end{subfigure}%
\begin{subfigure}[t]{0.22\textwidth}
  \centering
  \includegraphics[width=1.0\linewidth]{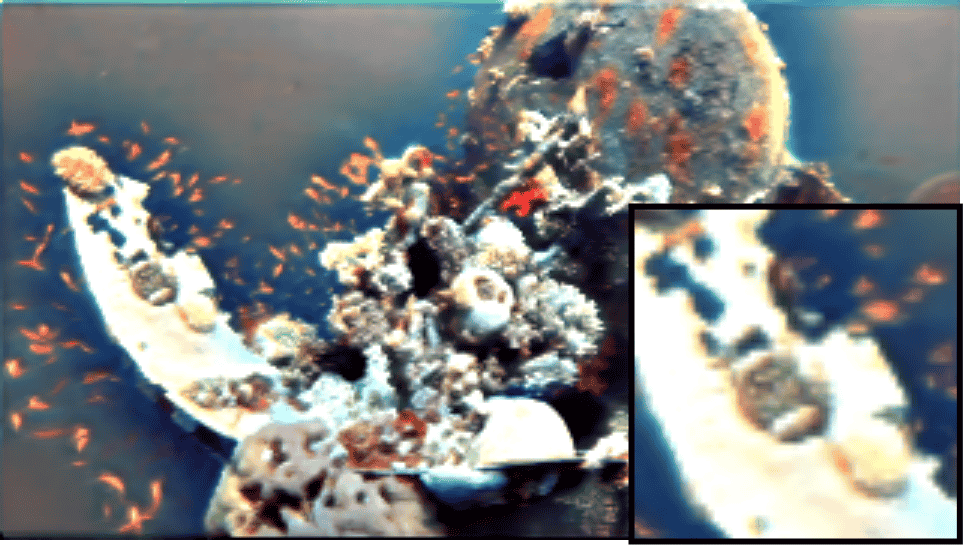}
\label{fig:starfish_uies} 
\end{subfigure}%
\begin{subfigure}[t]{0.22\textwidth}
  \centering
  \includegraphics[width=1.0\linewidth]{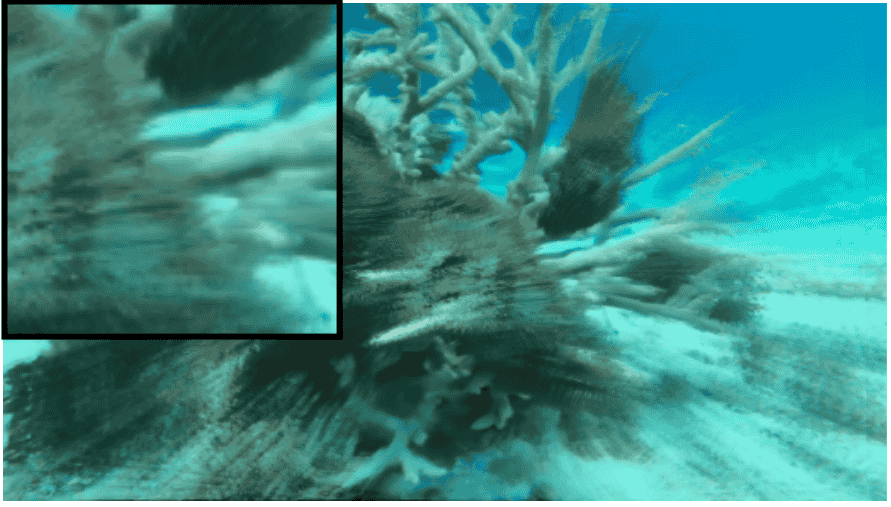}
\label{fig:starfish_upifm} 
\end{subfigure}%
\begin{subfigure}[t]{0.22\textwidth}
  \centering
  \includegraphics[width=1.0\linewidth, height=21.7mm]{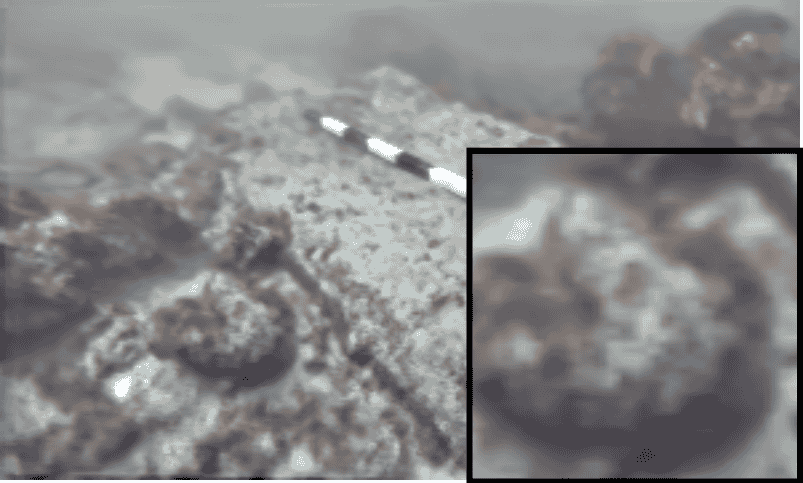}
\label{fig:starfish_nerf_clean} 
\end{subfigure}%

\setcounter{subfigure}{0}
\centering
\rotatebox{90}{\footnotesize U2NeRF(Gen+FT)}
\begin{subfigure}[t]{0.22\textwidth}
  \centering
  \includegraphics[width=1.0\linewidth]{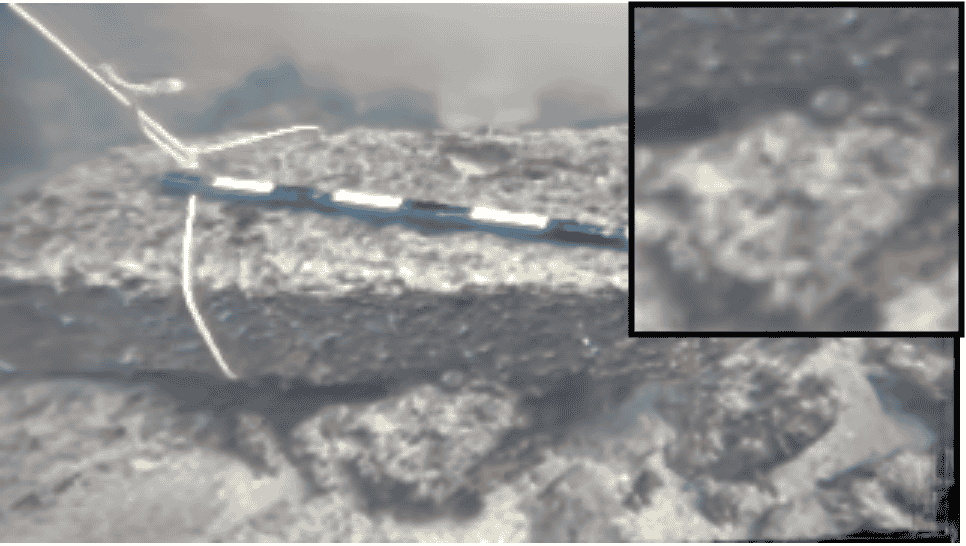}
\caption{Scene 1(Hard)}
\label{fig:dwarka26_uw} 
\end{subfigure}%
\begin{subfigure}[t]{0.22\textwidth}
  \centering
  \includegraphics[width=1.0\linewidth]{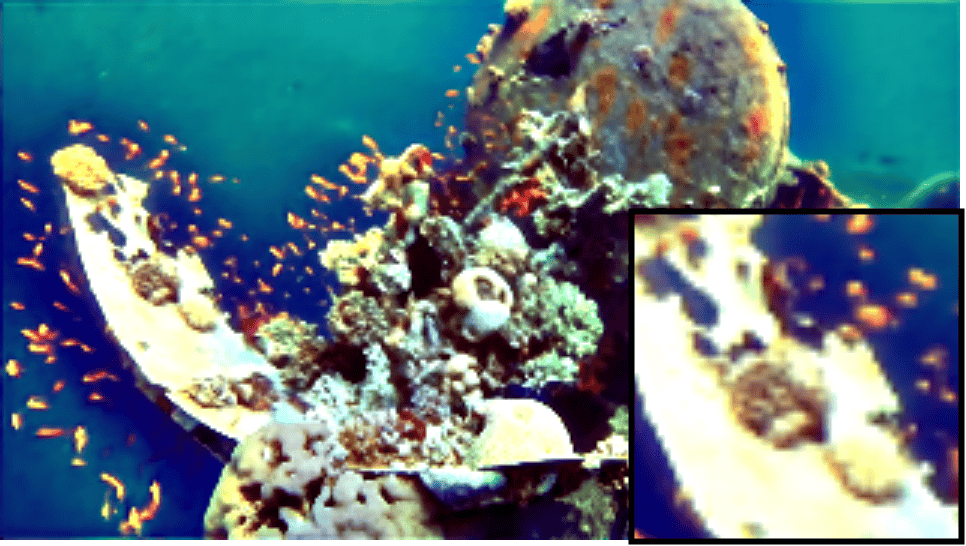}
\caption{Debris(Medium)}
\label{fig:dwarka26_uies} 
\end{subfigure}%
\begin{subfigure}[t]{0.22\textwidth}
  \centering
  \includegraphics[width=1.0\linewidth]{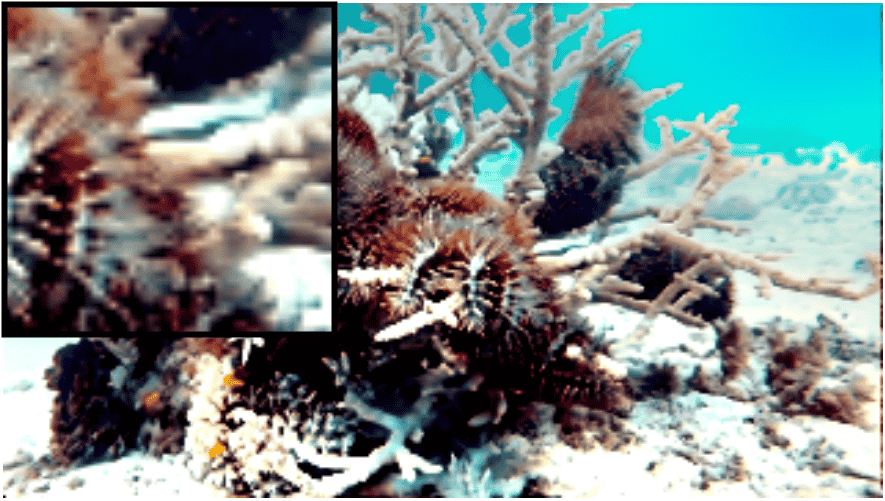}
\caption{Starfish(Medium)}
\label{fig:dwarka26_upifm} 
\end{subfigure}%
\begin{subfigure}[t]{0.22\textwidth}
  \centering
  \includegraphics[width=1.0\linewidth, height=21.7mm]{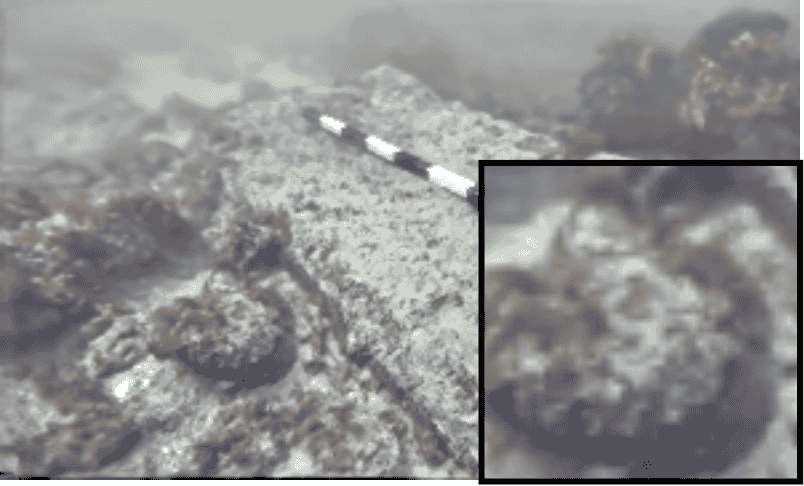}
\caption{Scene 4(Hard)}
\label{fig:dwarka26_nerf_clean} 
\end{subfigure}%
\caption{Qualitative results for cross-scene rendering. We visualize the underwater scene (row-1), novel views rendered using the pretrained network (row-2), novel views rendered using the finetuned network across different scenes (from left to right). U2NeRF successfully generalizes across scenes and when finetuned captures more intricate details. }
\label{fig:cross}
\vspace{-1.0em}
\end{figure*}

\subsection{Underwater View Synthesis Dataset}

\begin{figure}[ht]
\subfloat[Scene1 from hard split]{\label{a}\includegraphics[width=.45\linewidth]{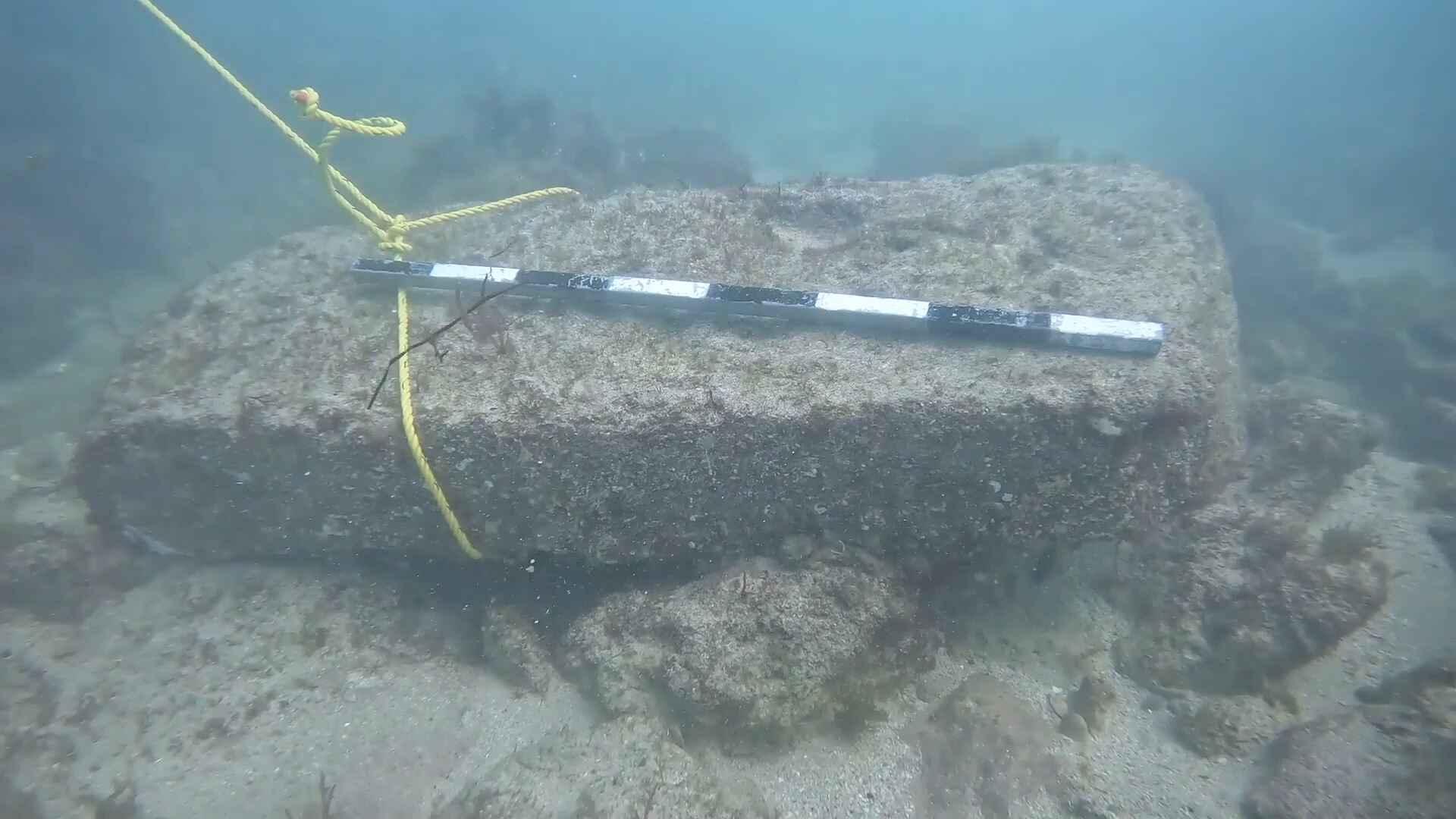}}\hfill
\subfloat[Coral from medium split]{\label{b}\includegraphics[width=.43\linewidth]{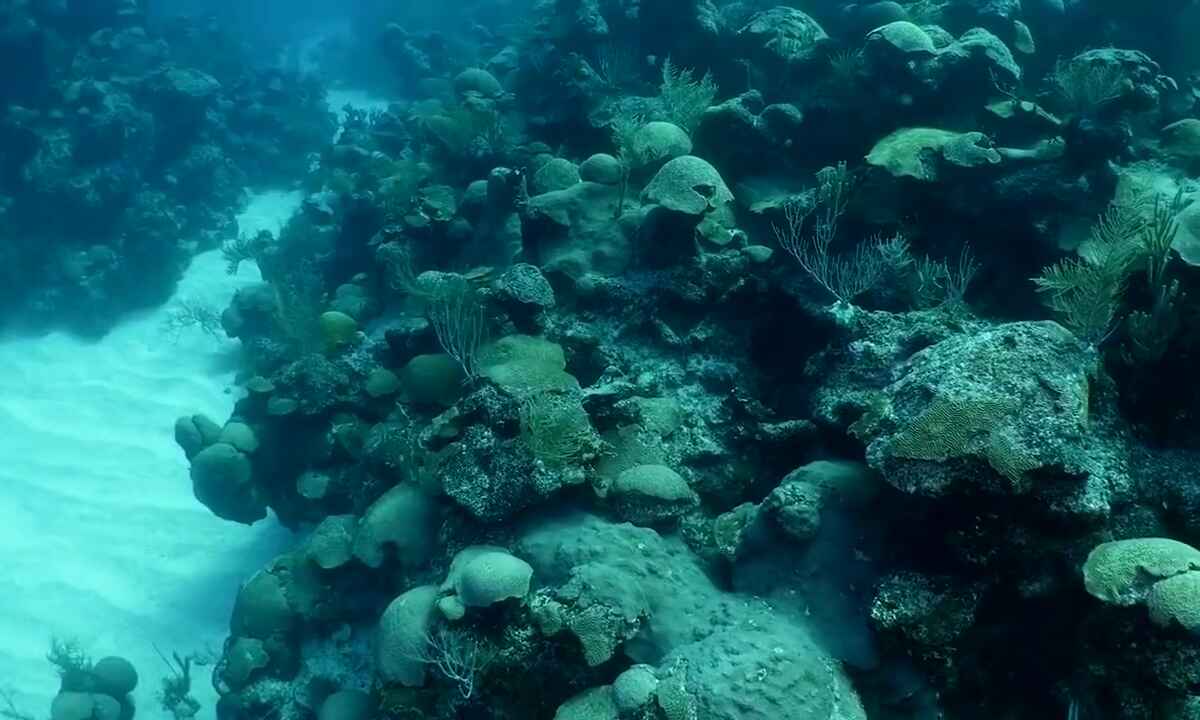}}\par 
\vspace{2mm}
\hspace{0.3\linewidth}\subfloat[Fern from easy split]{\label{c}\includegraphics[width=.35\linewidth]{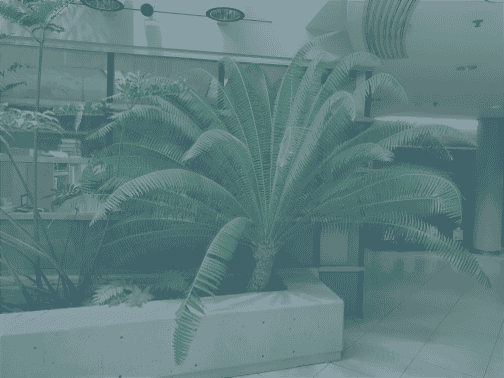}}
\caption{Illustrative examples of scenes from the UVS Dataset, one scene from each split is shown here.}
\label{fig:uvs}
\end{figure}

Due to the absence of multi-view underwater scene datasets suitable to evaluate novel-view rendering, we establish a new benchmark Underwater View Synthesis (UVS) Dataset containing 12 scenes, equally split into \textit{easy} (synthetic underwater scenes), \textit{medium} (real-world high quality), \textit{hard} (real-world low quality). The easy split contains 4 scenes from the LLFF dataset~\cite{mildenhall2019local}, namely ``fern'', ``fortress'', ``flower'', ``trex'' that were synthetically corrupted to simulate underwater scenes~\cite{desai2022aquagan}. For real-world data, we hand-pick 4 scenes from high quality youtube videos to form the medium split, while the hard split is composed of low-quality noisy real-world captures obtained during a diving expedition. For each scene from the medium and hard splits, we select roughly 100-150 images and calibrate them using Structure-from-Motion (SfM) algorithm in an open-source software package COLMAP \cite{schoenberger2016sfm, schoenberger2016mvs}. For COLMAP, we use a ``simple radial'' camera model with a single radial distortion coefficient and a shared intrinsic for all images. We use a ``sift feature guided matching'' option in the exhaustive matcher step of SfM and also refine principle points of the intrinsic during the bundle adjustment. Fig. \ref{fig:uvs} provides an illustration of the scenes present in the easy, medium, hard splits. 

\subsection{Baselines}
\label{sec:baselines}
The task of simultaneous restoration and rendering is novel and hence we establish several baselines to compare U2NeRF against in the UVS benchmark. We select NeRF as our neural renderer and identify different strategies to automatically ``restore'' the rendered views. As an initial baseline (labelled \textit{NeRF}), we train a vanilla NeRF model on the original underwater scenes, append state-of-the-art (SOTA) restoration methods as a post processing strategy (labelled \textit{NeRF + Clean}), and even train a NeRF model on restored images (labelled \textit{Clean + NeRF}). Additionally, we also consider non-rendering baselines where we assume access to the target view and attempt to restore it. We leverage SOTA underwater image restoration pipelines - UIESS~\cite{chen2022domain}, Underwater Physics-informed Image Formation Model (which we label \textit{UPIFM})~\cite{chai2022unsupervised}. 


\subsection{Single Scene Results}
\label{sec:single_scene_results}

\paragraph{Datasets. } To evaluate the single scene view generation capacity of U2NeRF, we perform experiments on the easy, medium and hard splits from the UVS dataset. We report average scores across all scenes within each split - easy: [Fern, Fortress, Flower, Trex], medium: [Starfish, Coral, Debris, Shipwreck], hard: [scene1, scene2, scene3, scene4] 
in Tables. \ref{tab:single_scene_easy}, \ref{tab:single_scene_medium}, \ref{tab:single_scene_hard} respectively.

\begin{table*}
\vspace{1em}
  \label{tab:single_scene}
  \centering
  \begin{subtable}[t]{0.45\textwidth}
  \centering
  \resizebox{0.9\columnwidth}{!}{
  \begin{tabular}{lccc}
    \toprule
    Models & UIQM$\uparrow$ & UCIQE$\uparrow$ & LPIPS (gray)$\downarrow$ \\
    \midrule
    U2NeRF Single Scene & 1.570 & 32.556 & 0.174\\
    U2NeRF Generalized & 1.426 & 32.293 & 0.279\\
    \midrule
    U2NeRF Generalized \\+ Finetuned & 1.856 & 34.113 & 0.222\\
    \bottomrule
  \end{tabular}
  }
  \caption{Medium Split}
  \label{tab:medium_split}
  \end{subtable}
  \begin{subtable}[t]{0.45\textwidth}
  \centering
  \resizebox{0.9\columnwidth}{!}{
  \begin{tabular}{lccc}
    \toprule
    Models & UIQM$\uparrow$ & UCIQE$\uparrow$ & LPIPS (gray)$\downarrow$ \\
    \midrule
    U2NeRF Single Scene & 1.100 & 26.788 & 0.260\\
    U2NeRF Generalized & 1.000 & 23.548 & 0.290\\
    \midrule
    U2NeRF (Generalized \\+ Finetuned) & 1.093 & 23.530 & 0.265\\
    \bottomrule
  \end{tabular}}
  \caption{Hard Split}
  \label{tab:llff_avg}
  \end{subtable}
   \caption{Comparison of U2NeRF (with and without fine-tuning) against baseline methods for cross-scene rendering on the UVS dataset}
   \label{tab:cross_scene}
\end{table*}

\paragraph{Discussion. } We compare U2NeRF against the baselines discussed in Sec. \ref{sec:baselines}. On the easy split, our proposed method achieves moderate PSNR scores but best LPIPS scores when compared to other rendering baselines by ~20\%. This could be because PSNR fails to measure structural distortions, blurring, has high sensitivity towards brightness, and hence does not effectively
measure visual quality. Similar inferences are discussed in~\cite{lin2023visionnerf} regarding discrepancies in
PSNR scores and their correlation to rendered image quality. In the case of more complex scenes present in the medium, hard data splits, we can clearly see the superiority of U2NeRF both in terms of rendering (LPIPS $\downarrow 11\%$), and color restoration quality (UIQM $\uparrow 5\%$, UCIQE $\uparrow 4\%$). 
More interestingly, we find that U2NeRF even outperforms no rendering baselines, that is, those algorithms that assume direct access to the target view and perform only restoration. Although our method extends upon UPIFM, we still manage to outperform the `only restoration' baseline with sufficient margin. This signifies the relevance of multi-view geometry to automatically restore a target view.  We show qualitative results in Fig. \ref{fig:single}, and can clearly see that U2NeRF renders and restores images with greater visual quality when compared to other methods. In the case of debris, U2NeRF successfully recovers the fishes and enhances its visibility to improve restoration quality, while in the case of scene2, U2NeRF is able to render complex, moving structures like ropes while still maintaining higher detail along the surface of the rock.

\subsection{Cross Scene Results}

\paragraph{Datasets. } U2NeRF leverages multi-view features complying with epipolar geometry, enabling generalization to unseen scenes. We randomly select scenes from video data captured during our diving expedition, and we use a total of 45 scenes for training. Table. \ref{tab:cross_scene} discusses results of the trained network on all 8 scenes from the medium and hard splits in the UVS dataset. Please note that the scenes from the UVS dataset are held out during training to gauge the model's generalization performance. 

\paragraph{Discussion. } We compare U2NeRF's generalization performance with a corresponding network trained only on a single scene. Although the network is only trained on data captured during our diving expedition, it still manages to generalize to unseen objects present in the medium split. Once finetuned (with as little as 50k training steps), U2NeRF manages to perform as well or even outperform a vanilla NeRF trained on each scene. Fig. \ref{fig:cross} visualizes qualitative results on the UVS dataset. We can clearly see that the pre-trained model can successfully generalize across several scenes and when finetuned, further improves performance.

\subsection{Ablation Studies}

\begin{table}[ht]
  \centering
  \resizebox{0.8\columnwidth}{!}{
  \begin{tabular}{lccc}
    \toprule
    Models & UIQM$\uparrow$ & UCIQE$\uparrow$ & LPIPS (gray)$\downarrow$ \\
    \midrule
    U2NeRF ($p$ = 2) & 1.964 & 34.234 & 0.249\\
    U2NeRF ($p$ = 4) & 2.222 & 35.120 & 0.187\\
    U2NeRF ($p$ = 8) & 2.096 & 34.011 & 0.202\\
    \bottomrule
   
  \end{tabular}}
  \caption{Effect of Patch Size}
  \label{tab:patch}
\end{table}

\begin{table}[ht]
  \centering
  \resizebox{0.8\columnwidth}{!}{
  \begin{tabular}{lccc}
    \toprule
    Models & UIQM$\uparrow$ & UCIQE$\uparrow$ & LPIPS (gray)$\downarrow$ \\
    \midrule
    U2NeRF ($N$ = 3) & 2.208 & 35.084 & 0.191\\
    U2NeRF ($N$ = 5) & 2.214 & 35.106 & 0.189\\
    U2NeRF ($N$ = 8) & 2.220 & 35.116 & 0.188\\
    U2NeRF ($N$ = 10) & 2.222 & 35.120 & 0.187\\
    \bottomrule
  \end{tabular}}
  \caption{Effect of number of source views}
  \label{tab:src_views}
\end{table}

\paragraph{Effect of patch size. } To verify the effect of patch size on the rendered image quality, we train U2NeRF on the starfish scene with varying patch sizes (2, 4, and 8). From Table. \ref{tab:patch}, we can clearly see that $p$ = 4 yields the best results across all three metrics. A larger patch size requires more information (beyond just epipolar points), for accurate reconstruction, while a smaller patch size does not act as a useful prior for restoration. Therefore, patch size = 4 strikes the ideal balance between performance and network complexity. 

\begin{figure}[ht]
\centering
\begin{subfigure}[t]{0.24\textwidth}
  \centering
  \includegraphics[width=1.0\linewidth]{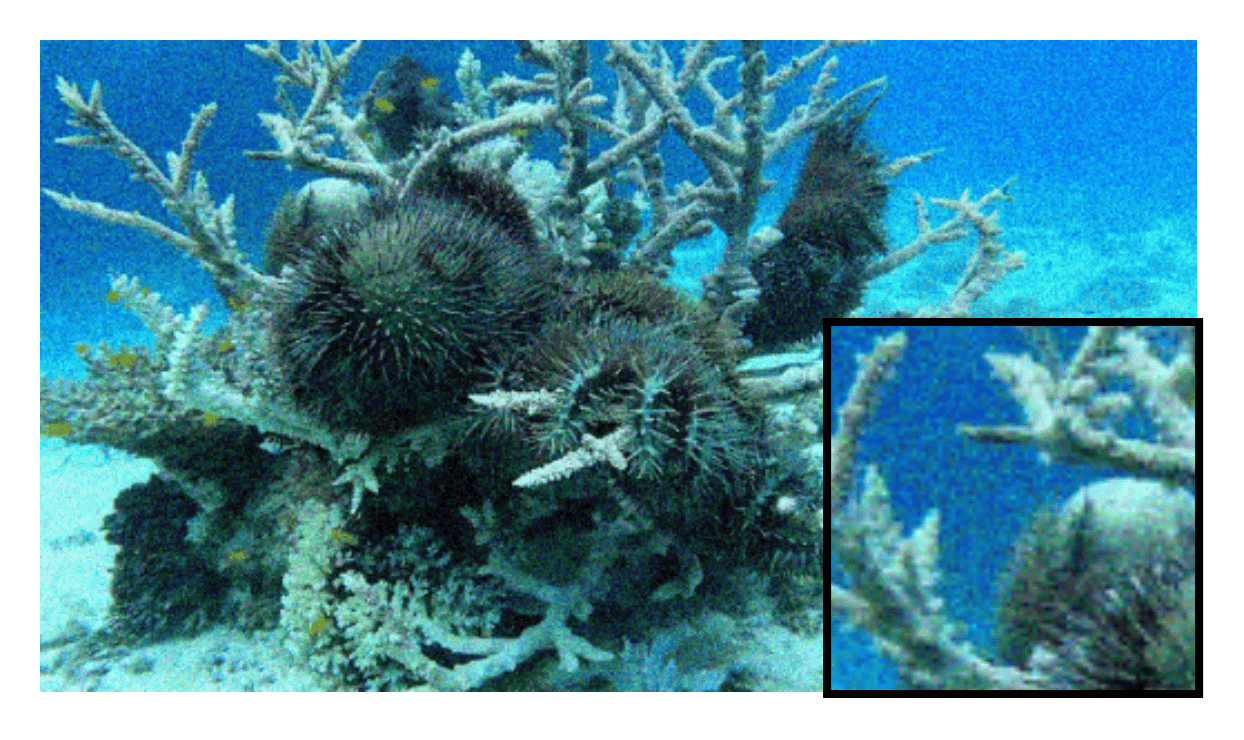}
\label{fig:starfish_corrupted} 
\end{subfigure}%
\begin{subfigure}[t]{0.24\textwidth}
  \centering
  \includegraphics[width=1.0\linewidth]{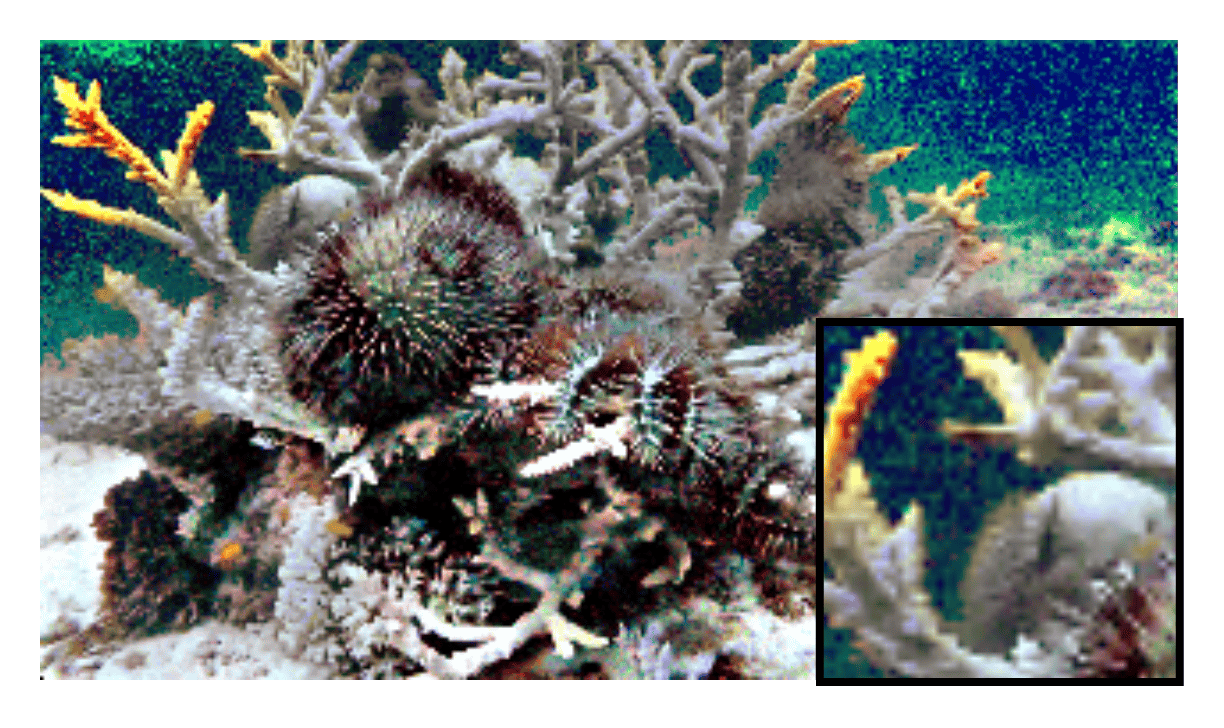}
\label{fig:starfish_denoised} 
\end{subfigure}%
\setcounter{subfigure}{0}
\begin{subfigure}[t]{0.24\textwidth}
  \centering
  \includegraphics[width=1.0\linewidth]{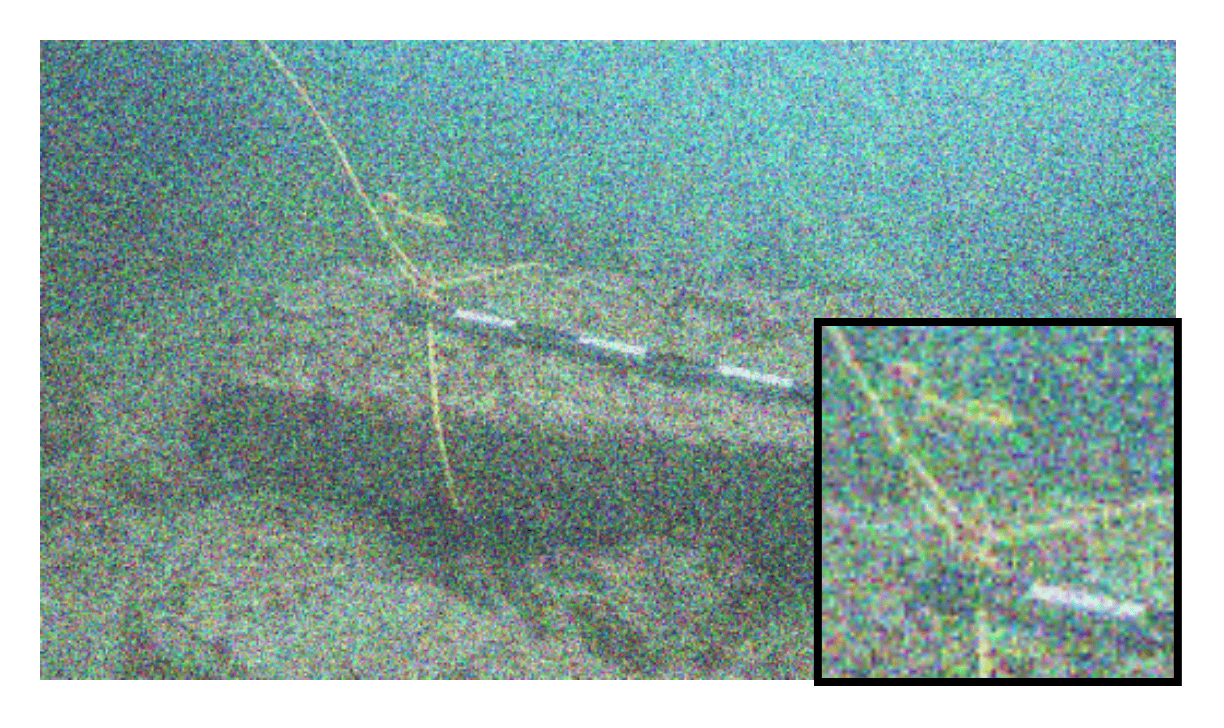}
\caption{Underwater Image}
\label{fig:dwarka1_corrupted} 
\end{subfigure}%
\begin{subfigure}[t]{0.24\textwidth}
  \centering
  \includegraphics[width=1.0\linewidth]{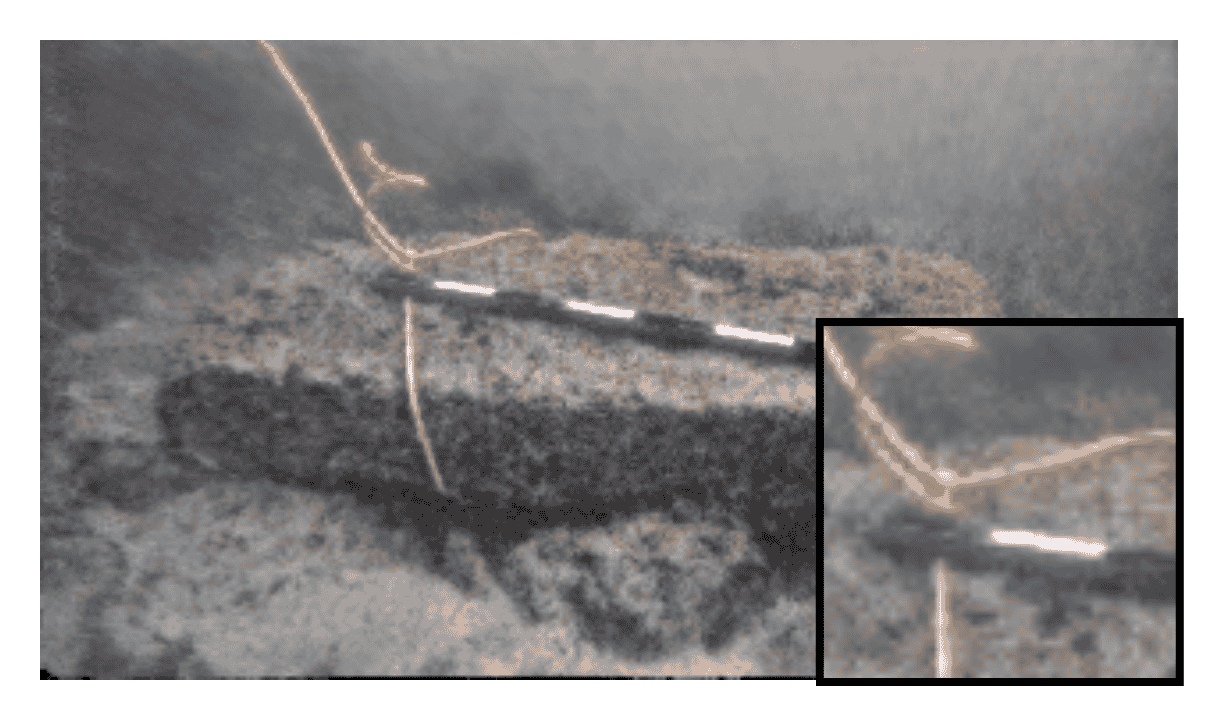}
\caption{U2NeRF}
\label{fig:dwarka1_denoised} 
\end{subfigure}%
\caption{Denoising results of U2NeRF}
\label{fig:corruption}
\vspace{-1.0em}
\end{figure}


\begin{figure*}[ht]
\centering
\begin{subfigure}[t]{0.18\textwidth}
  \centering
  \includegraphics[width=1.0\linewidth]{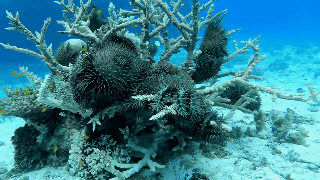}
\label{fig:debris_uw} 
\end{subfigure}%
\begin{subfigure}[t]{0.18\textwidth}
  \centering
  \includegraphics[width=1.0\linewidth]{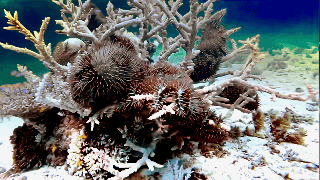}
\label{fig:debris_uies} 
\end{subfigure}%
\begin{subfigure}[t]{0.18\textwidth}
  \centering
  \includegraphics[width=1.0\linewidth]{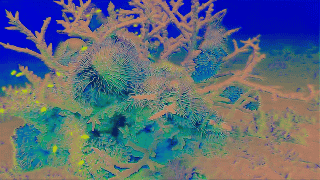}
\label{fig:debris_upifm} 
\end{subfigure}%
\begin{subfigure}[t]{0.18\textwidth}
  \centering
  \includegraphics[width=1.0\linewidth]{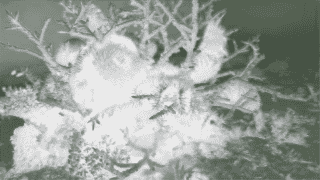}
\label{fig:debris_nerf_clean} 
\end{subfigure}%
\begin{subfigure}[t]{0.18\textwidth}
  \centering
  \includegraphics[width=1.0\linewidth]{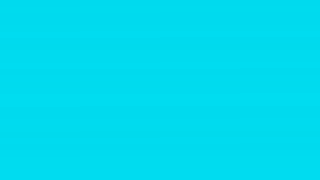}
\label{fig:debris_u2nerf} 
\end{subfigure}%
\setcounter{subfigure}{0}
\centering
\begin{subfigure}[t]{0.18\textwidth}
  \centering
  \includegraphics[width=1.0\linewidth]{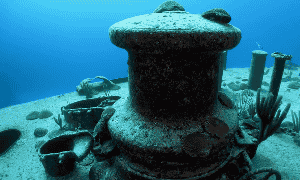}
\caption{\footnotesize{Degraded Image}}
\label{fig:starfish_uw} 
\end{subfigure}%
\begin{subfigure}[t]{0.18\textwidth}
  \centering
  \includegraphics[width=1.0\linewidth]{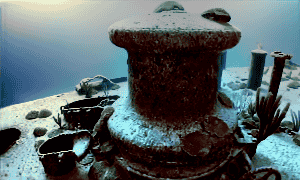}
\caption{Restored Image}
\label{fig:starfish_uies} 
\end{subfigure}%
\begin{subfigure}[t]{0.18\textwidth}
  \centering
  \includegraphics[width=1.0\linewidth]{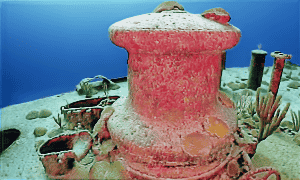}
\caption{\footnotesize{Direct Transmission}}
\label{fig:starfish_upifm} 
\end{subfigure}%
\begin{subfigure}[t]{0.18\textwidth}
  \centering
  \includegraphics[width=1.0\linewidth]{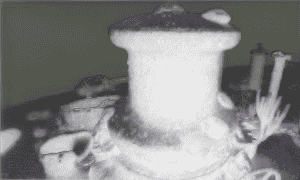}
\caption{\footnotesize{Backscatter Trans.}}
\label{fig:starfish_nerf_clean} 
\end{subfigure}%
\begin{subfigure}[t]{0.18\textwidth}
  \centering
  \includegraphics[width=1.0\linewidth]{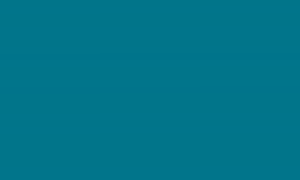}
\caption{\footnotesize{Global Light}}
\label{fig:starfish_u2nerf} 
\end{subfigure}%
\caption{Visualisations of the predicted image components (scene radiance, transmission maps, global light).}
\label{fig:maps}
\vspace{-1.0em}
\end{figure*}

\paragraph{Effect of sparse source views. } To test the performance of U2NeRF in the presence of sparse views, we evaluate a trained model on the starfish scene but with fewer source views. From Table. \ref{tab:src_views}, we can see that as the number of source views increase, the model performs better. However, there is almost no significant drop in performance even when only 3 source views are given as input. This verifies the suitability of U2NeRF even in the presence of sparse input views. 

\paragraph{Effect of Gaussian noise. } Unlike standard NeRF methods, U2NeRF predicts an image patch rather than a single pixel color. Therefore, we hypothesize that without explicit training, U2NeRF can denoise small perturbations present in the scene. To verify this claim, we evaluate a trained U2NeRF model on scenes corrupted using a Gaussian noise (unseen) with mean 0, and standard deviation 0.05. We show qualitative results in Fig \ref{fig:corruption}.

\subsection{Physical Interpretation of U2NeRF}

U2NeRF attempts to implicitly predict individual image components $J$, $T_{D}$, $T_{B}$ and $A$ which when combined restores the underwater image. Fig \ref{fig:maps} visualizes examples of underwater scenes with their corresponding predicted image components. We can clearly see that visualized $T_{D}$ and $T_{B}$ maps closely simulate depth which is consistent with the physics-informed image formation model~\cite{chai2022unsupervised}. $A$ indicates the global background light which corresponds to the brightest pixel in the scene which can be confirmed from Fig \ref{fig:maps}. The restored image in Fig \ref{fig:maps} corresponds to the scene radiance. Therefore, with no explicit supervision, U2NeRF learns to physically ground its learnable operations. 

\section{Conclusion}

We present Unsupervised Underwater NeRF (U2NeRF), that extends radiance fields to simultaneously render and restore novel views, more specifically in the context of underwater images. We demonstrate that by augmenting existing radiance fields with spatial awareness, and when combined with a physics-informed underwater image formation model can successfully restore underwater images. Additionally, we contribute a novel Underwater View Synthesis Dataset (UVSDataset) consisting of 12 underwater scenes, containing both synthetically generated, and real-world data. Extensive experiments reveal that U2NeRF outperforms existing baselines and achieves best perceptual metric scores (LPIPS $\downarrow$ 11\%, UIQM $\uparrow$ 5\%, UCIQE $\uparrow$ 4\%). These results demonstrate that transformers can be successfully used to model the underlying physics in 3D vision. 

{\small
\bibliographystyle{ieee_fullname}
\bibliography{main}
}

\clearpage
\appendix

\section{Loss Functions}
\label{sec:intro}

To achieve the rendering and restoration in an unsupervised manner, it is important to regularise the model with appropriate losses. We propose 6 distinct loss functions which act upon the different output maps similar to \cite{chai2022unsupervised}

\subsection{Reconstruction Loss} 
We use reconstruction loss to self-supervise the layer decomposition process. We supervise it using an MSE loss between the original underwater picture and the predicted image. 
We aim to minimize the loss $\mathcal{L}_{Rec}$ as below
\begin{equation} \label{eu_eqn}
    \mathcal{L}_{Rec} = || I - \Mat{x} ||_2^2
\end{equation}
where I is the ground truth image and x represents the predicted image. 

\subsection{Contrast Enhancement Loss} 
The difference between brightness and saturation is almost zero in a clean image as observed in \cite{zhu2015fast}. The contrast enhancement loss $\mathcal{L}_{Con}$ is created as follows to supervise scene radiance map($J$):
\begin{equation} \label{eu_eqn}
    \mathcal{L}_{Con} = || V(J(\Mat{x})) - S(J(\Mat{x})) ||_2^2
\end{equation}
where V represents brightness and S represents saturation of scene radiance $J(\Mat{x})$

\subsection{Color Constancy Loss} 
To rectify any potential colour inconsistencies in the recovered image, we build a colour constancy loss in line with the Gray-World colour constancy theory \cite{buchsbaum1980spatial}. $\mathcal{L}_{Col}$ describes the loss as follows:
\begin{equation} \label{eu_eqn}
    \mathcal{L}_{Col} = \sum_{c \in \Omega}|| \mu(J_c) - 0.5 ||_2^2, \Omega = {R, G, B}
\end{equation}
where $\mu(J_c)$ represents the average intensity value of color channel c in the estimated scene radiance.

\subsection{Light Global Property Loss} In order to reduce the discrepancy between the latent code z and the reconstruction of the latent code $\hat{z}$ in the A-Net, light global property loss $\mathcal{L}_{kl}$ is created for variational inference.
\begin{equation} \label{eu_eqn}
    \mathcal{L}_{kl} = KL(\mathcal{N}(\mu_z,\sigma_z^2) || \mathcal{N}(0,I))
\end{equation}
where KL(·) denotes the Kullback-Leibler divergence be- tween two distributions, $\mathcal{N}(\mu_z,\sigma_z^2)$ denotes the learned latent Gaussian distribution, and $\mathcal{N}$(0, I) refers to the standard normal distribution.

\subsection{Transmission Consistency Loss} Since the backscatter co-efficient solely depend on the optical properties of the water, they should be constants in the backscatter transmission map. We propose a transmission consistency loss to supervise backscatter transmission map($T_{B}$). According to the loss $\mathcal{L}_{T}$,
\begin{equation} \label{eu_eqn}
    \mathcal{L}_{T} = \sum_{c_1, c_2 \in \epsilon} || \frac{\log{T^c_1}}{\log{T^c_2}} - \mu(\frac{\log{T^c_1}}{\log{T^c_2}}) ||_2^2
\end{equation}
where $T_c$ stands for the estimated backscatter transmission map of the c channel, $\mu$ is the average factor, and $\epsilon = \{(R, G), (R, B), (G, B)\}$ is a collection of colour pairs.

\subsection{Global Consistency Loss} 
The goal of global consistency loss $\mathcal{L}_{Glob}$ is to blur/smoothen the global background light($A$) completely. This loss indirectly enforces the smoothness criteria by requiring each pixel to have the same colour as the adjacent pixel.

\vspace{3mm}

\textbf{Total Loss} The total loss of our method is as below:
\begin{equation*} \label{eu_eqn}
\begin{aligned}
    \mathcal{L} = \lambda_1\mathcal{L}_{Rec}, + \lambda_2\mathcal{L}_{Con} + \lambda_3\mathcal{L}_{Col} + \lambda_4\mathcal{L}_{kl}\\  +  \lambda_5\mathcal{L}_{T} + \lambda_6\mathcal{L}_{Glob} 
\end{aligned}
\end{equation*}

where $\lambda$ is the weight. We set $\lambda_1$ = 1,$\lambda_2$ = 0.1, $\lambda_3$ = 1, $\lambda_4$ = 1, $\lambda_5$ = 0.1, $\lambda_6$ = 1 to obtain the best results. 

\section{Results}
\begin{figure*}[ht!]
\centering


\begin{subfigure}[t]{0.15\textwidth}
  \centering
  \includegraphics[width=1.0\linewidth]{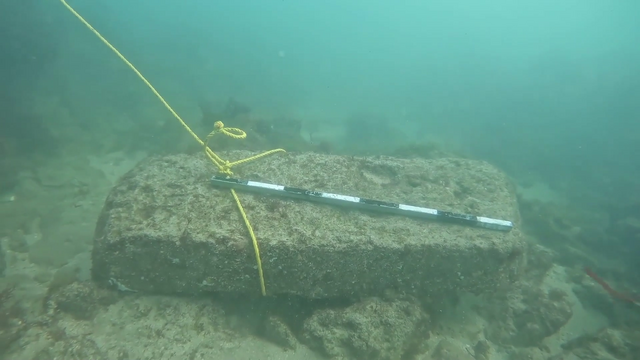}
\label{fig:debris_uw} 
\end{subfigure}%
\begin{subfigure}[t]{0.15\textwidth}
  \centering
  \includegraphics[width=1.0\linewidth]{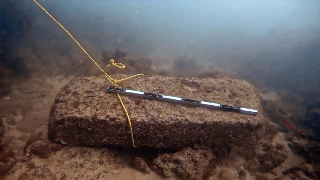}
\label{fig:debris_u2nerf} 
\end{subfigure}%
\begin{subfigure}[t]{0.158\textwidth}
  \centering
  \includegraphics[width=1.0\linewidth]{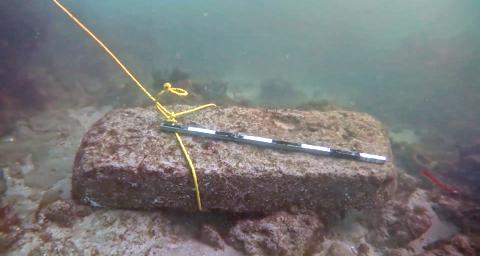}
\label{fig:debris_u2nerf} 
\end{subfigure}%
\begin{subfigure}[t]{0.15\textwidth}
  \centering
  \includegraphics[width=1.0\linewidth]{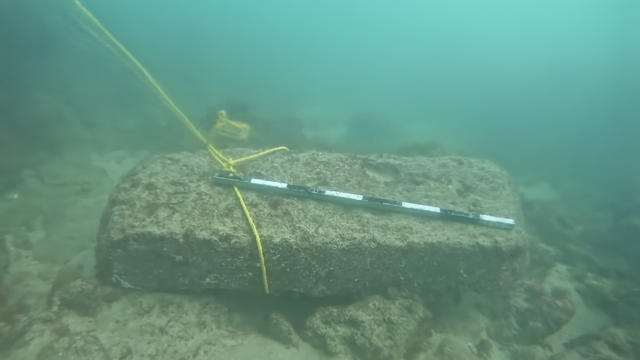}
\label{fig:debris_uw} 
\end{subfigure}%
\begin{subfigure}[t]{0.15\textwidth}
  \centering
  \includegraphics[width=1.0\linewidth]{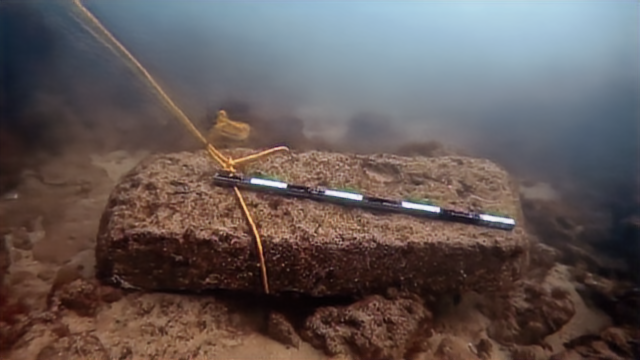}
\label{fig:debris_uies} 
\end{subfigure}%
\begin{subfigure}[t]{0.15\textwidth}
  \centering
  \includegraphics[width=1.0\linewidth]{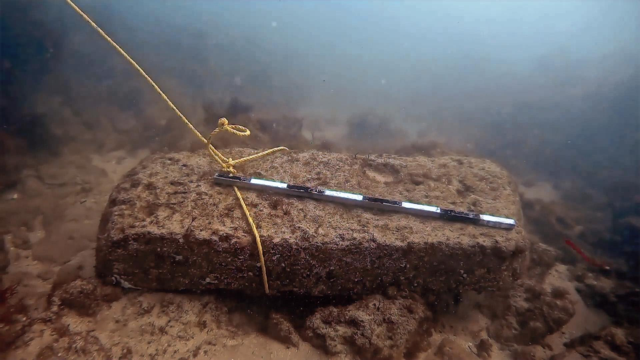}
\label{fig:debris_upifm} 
\end{subfigure}%
\begin{subfigure}[t]{0.15\textwidth}
  \centering
  \includegraphics[width=1.0\linewidth]{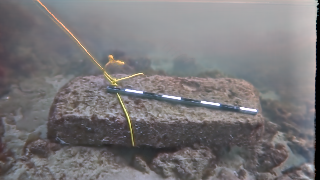}
\label{fig:debris_nerf_clean} 
\end{subfigure}%

\setcounter{subfigure}{0}
\centering

\begin{subfigure}[t]{0.15\textwidth}
  \centering
  \includegraphics[width=1.0\linewidth]{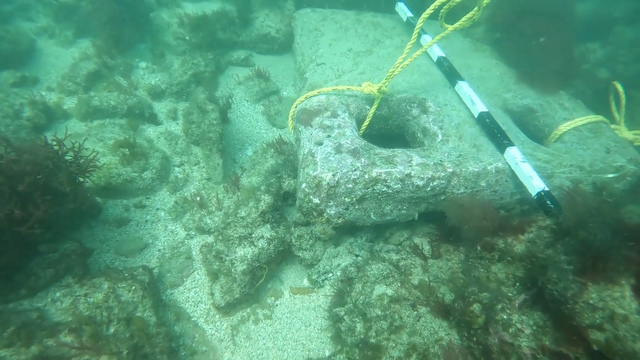}
\label{fig:debris_uw} 
\end{subfigure}%
\begin{subfigure}[t]{0.15\textwidth}
  \centering
  \includegraphics[width=1.0\linewidth]{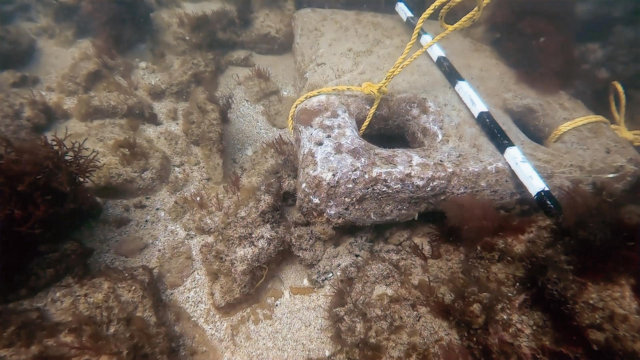}
\label{fig:debris_u2nerf} 
\end{subfigure}%
\begin{subfigure}[t]{0.158\textwidth}
  \centering
  \includegraphics[width=1.0\linewidth]{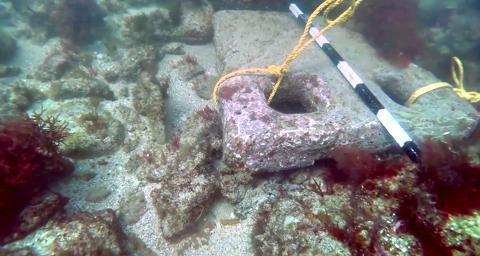}
\label{fig:debris_u2nerf} 
\end{subfigure}%
\begin{subfigure}[t]{0.15\textwidth}
  \centering
  \includegraphics[width=1.0\linewidth]{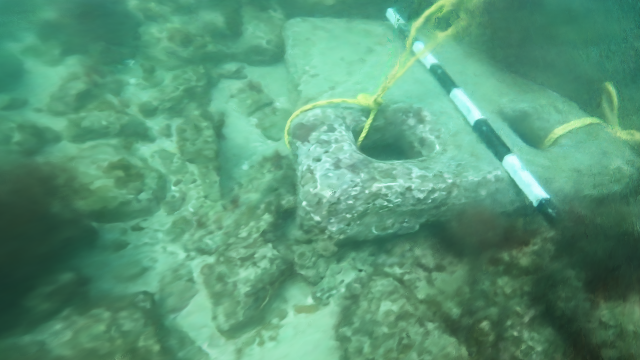}
\label{fig:debris_uw} 
\end{subfigure}%
\begin{subfigure}[t]{0.15\textwidth}
  \centering
  \includegraphics[width=1.0\linewidth]{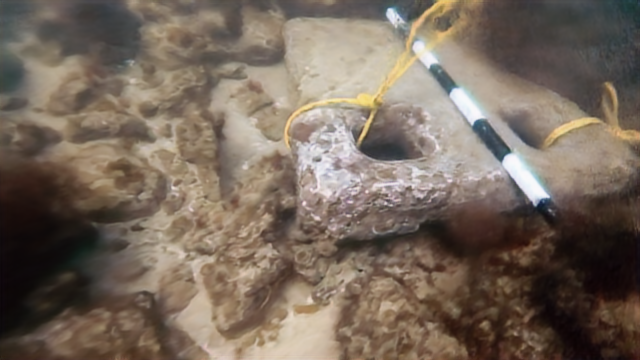}
\label{fig:debris_uies} 
\end{subfigure}%
\begin{subfigure}[t]{0.15\textwidth}
  \centering
  \includegraphics[width=1.0\linewidth]{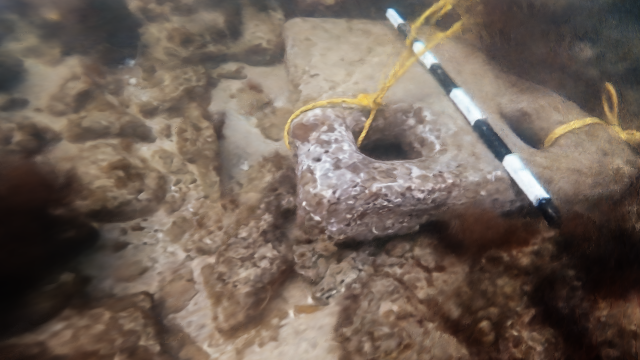}
\label{fig:debris_upifm} 
\end{subfigure}%
\begin{subfigure}[t]{0.15\textwidth}
  \centering
  \includegraphics[width=1.0\linewidth]{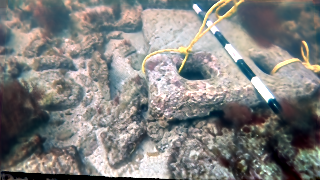}
\label{fig:debris_nerf_clean} 
\end{subfigure}%

\setcounter{subfigure}{0}
\centering

\begin{subfigure}[t]{0.15\textwidth}
  \centering
  \includegraphics[width=1.0\linewidth]{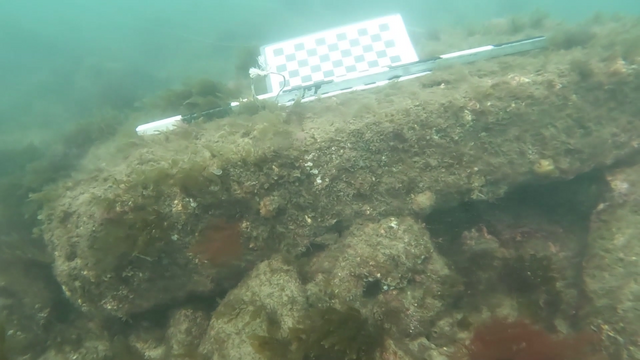}
\label{fig:debris_uw} 
\end{subfigure}%
\begin{subfigure}[t]{0.15\textwidth}
  \centering
  \includegraphics[width=1.0\linewidth]{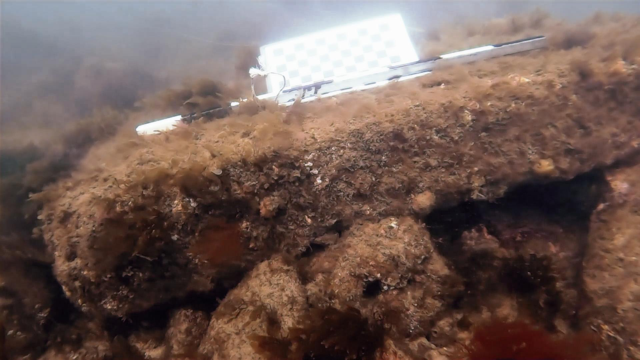}
\label{fig:debris_u2nerf} 
\end{subfigure}%
\begin{subfigure}[t]{0.158\textwidth}
  \centering
  \includegraphics[width=1.0\linewidth]{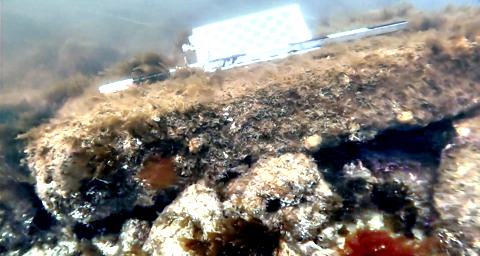}
\label{fig:debris_u2nerf} 
\end{subfigure}%
\begin{subfigure}[t]{0.15\textwidth}
  \centering
  \includegraphics[width=1.0\linewidth]{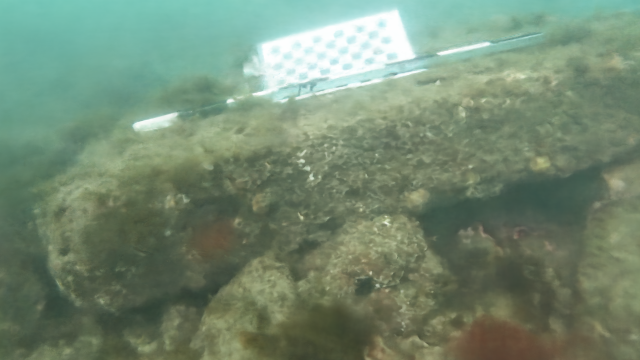}
\label{fig:debris_uw} 
\end{subfigure}%
\begin{subfigure}[t]{0.15\textwidth}
  \centering
  \includegraphics[width=1.0\linewidth]{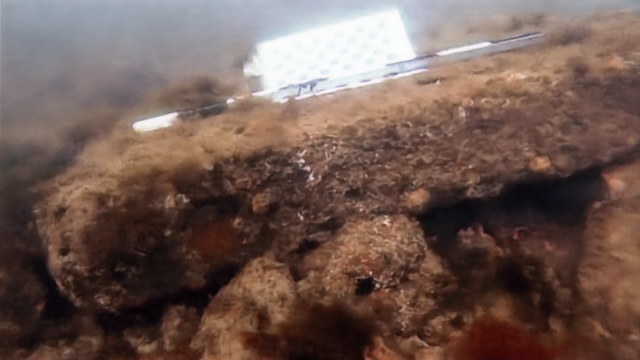}
\label{fig:debris_uies} 
\end{subfigure}%
\begin{subfigure}[t]{0.15\textwidth}
  \centering
  \includegraphics[width=1.0\linewidth]{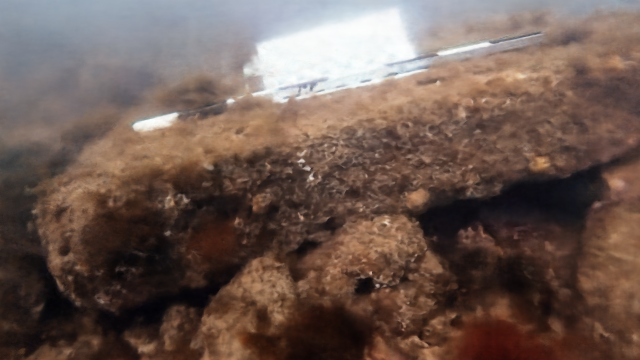}
\label{fig:debris_upifm} 
\end{subfigure}%
\begin{subfigure}[t]{0.15\textwidth}
  \centering
  \includegraphics[width=1.0\linewidth]{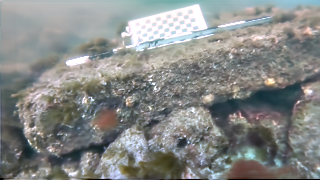}
\label{fig:debris_nerf_clean} 
\end{subfigure}%

\setcounter{subfigure}{0}
\centering

\begin{subfigure}[t]{0.15\textwidth}
  \centering
  \includegraphics[width=1.0\linewidth,height=0.6\linewidth]{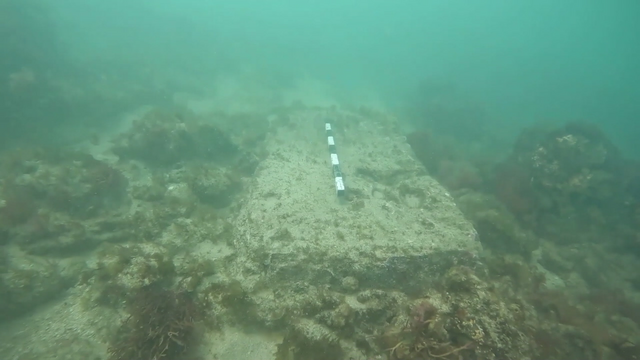}
\label{fig:debris_uw} 
\end{subfigure}%
\begin{subfigure}[t]{0.15\textwidth}
  \centering
  \includegraphics[width=1.0\linewidth]{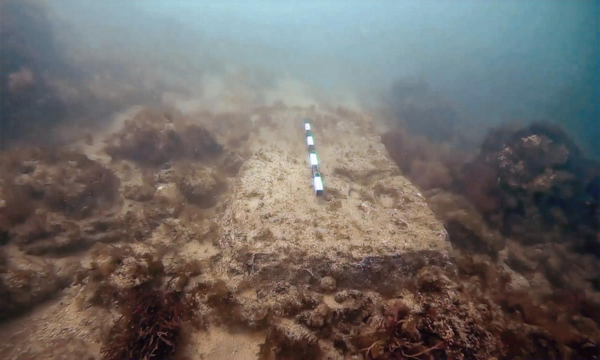}
\label{fig:debris_u2nerf} 
\end{subfigure}%
\begin{subfigure}[t]{0.158\textwidth}
  \centering
  \includegraphics[width=1.0\linewidth,height=0.57\linewidth]{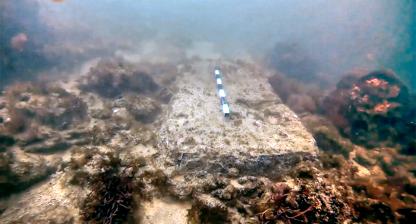}
\label{fig:debris_u2nerf} 
\end{subfigure}%
\begin{subfigure}[t]{0.15\textwidth}
  \centering
  \includegraphics[width=1.0\linewidth]{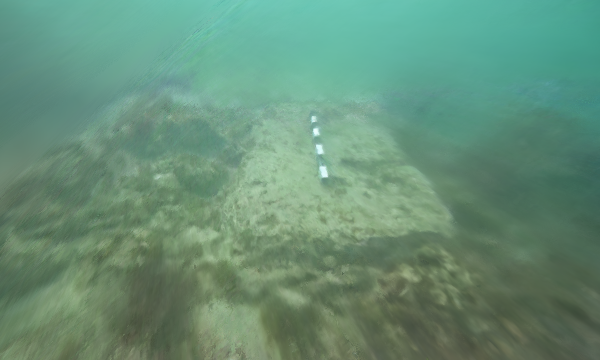}
\label{fig:debris_uw} 
\end{subfigure}%
\begin{subfigure}[t]{0.15\textwidth}
  \centering
  \includegraphics[width=1.0\linewidth]{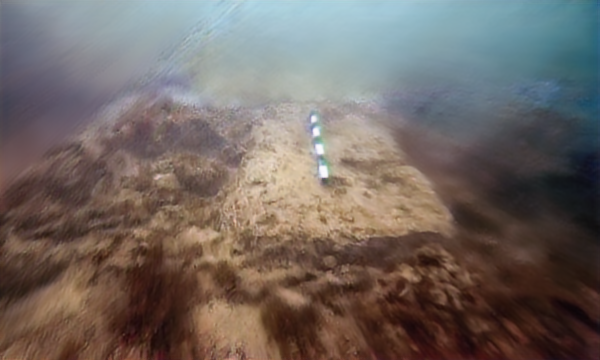}
\label{fig:debris_uies} 
\end{subfigure}%
\begin{subfigure}[t]{0.15\textwidth}
  \centering
  \includegraphics[width=1.0\linewidth,height=0.6\linewidth]{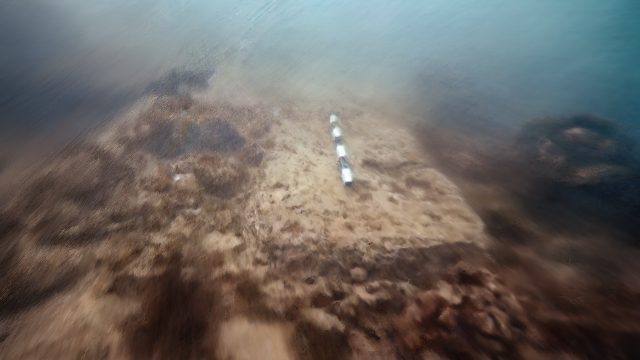}
\label{fig:debris_upifm} 
\end{subfigure}%
\begin{subfigure}[t]{0.15\textwidth}
  \centering
  \includegraphics[width=1.0\linewidth]{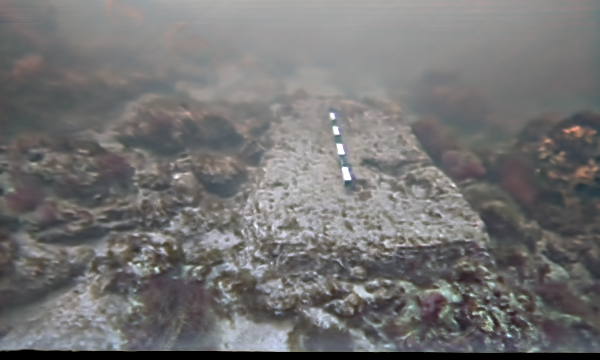}
\label{fig:debris_nerf_clean} 
\end{subfigure}%

\setcounter{subfigure}{0}
\centering

\begin{subfigure}[t]{0.15\textwidth}
  \centering
  \includegraphics[width=1.0\linewidth]{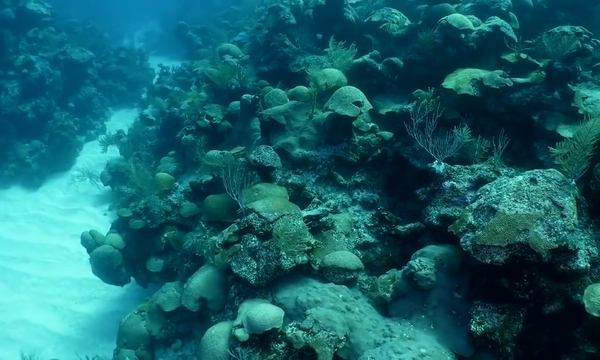}
\label{fig:debris_uw} 
\end{subfigure}%
\begin{subfigure}[t]{0.15\textwidth}
  \centering
  \includegraphics[width=1.0\linewidth]{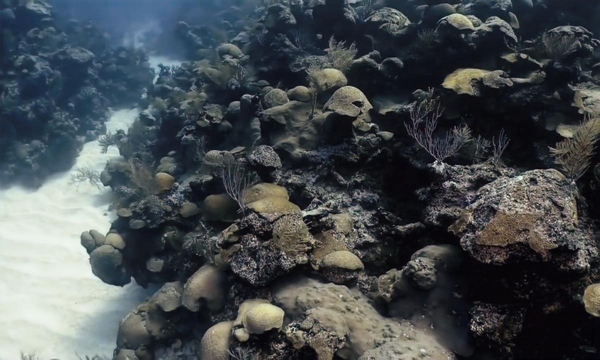}
\label{fig:debris_u2nerf} 
\end{subfigure}%
\begin{subfigure}[t]{0.158\textwidth}
  \centering
  \includegraphics[width=1.0\linewidth,height=0.57\linewidth]{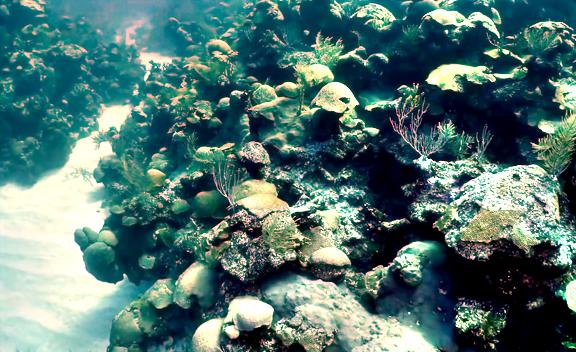}
\label{fig:debris_u2nerf} 
\end{subfigure}%
\begin{subfigure}[t]{0.15\textwidth}
  \centering
  \includegraphics[width=1.0\linewidth]{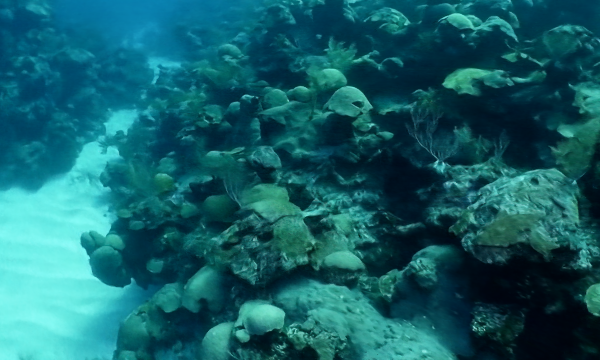}
\label{fig:debris_uw} 
\end{subfigure}%
\begin{subfigure}[t]{0.15\textwidth}
  \centering
  \includegraphics[width=1.0\linewidth]{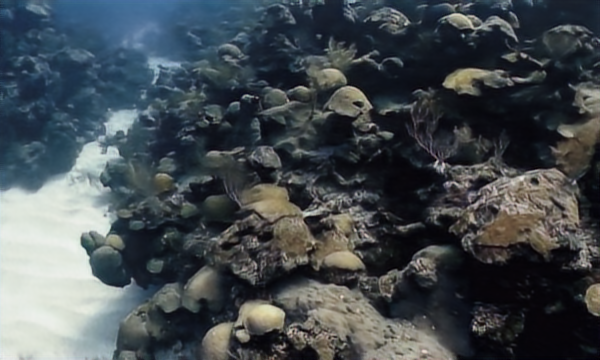}
\label{fig:debris_uies} 
\end{subfigure}%
\begin{subfigure}[t]{0.15\textwidth}
  \centering
  \includegraphics[width=1.0\linewidth]{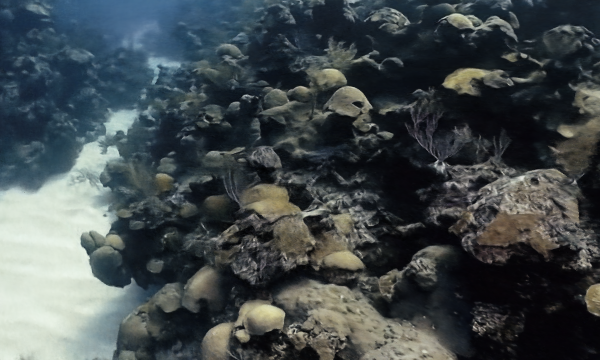}
\label{fig:debris_upifm} 
\end{subfigure}%
\begin{subfigure}[t]{0.15\textwidth}
  \centering
  \includegraphics[width=1.0\linewidth]{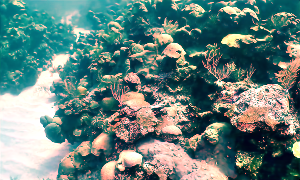}
\label{fig:debris_nerf_clean} 
\end{subfigure}%

\setcounter{subfigure}{0}
\centering

\begin{subfigure}[t]{0.15\textwidth}
  \centering
  \includegraphics[width=1.0\linewidth]{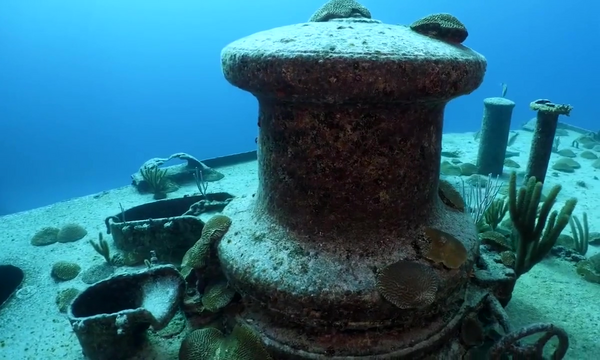}
\label{fig:debris_uw} 
\end{subfigure}%
\begin{subfigure}[t]{0.15\textwidth}
  \centering
  \includegraphics[width=1.0\linewidth]{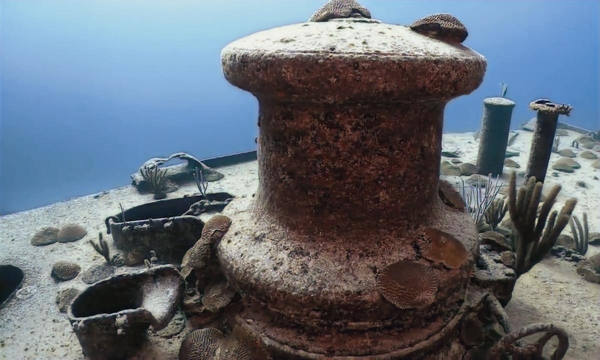}
\label{fig:debris_u2nerf} 
\end{subfigure}%
\begin{subfigure}[t]{0.158\textwidth}
  \centering
  \includegraphics[width=1.0\linewidth,height=0.57\linewidth]{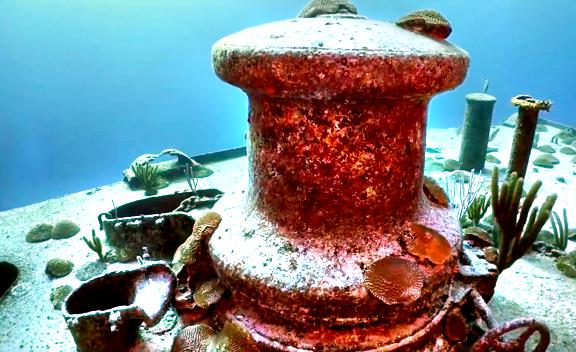}
\label{fig:debris_u2nerf} 
\end{subfigure}%
\begin{subfigure}[t]{0.15\textwidth}
  \centering
  \includegraphics[width=1.0\linewidth]{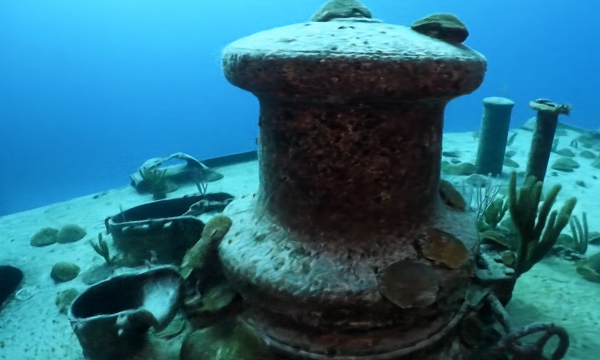}
\label{fig:debris_uw} 
\end{subfigure}%
\begin{subfigure}[t]{0.15\textwidth}
  \centering
  \includegraphics[width=1.0\linewidth]{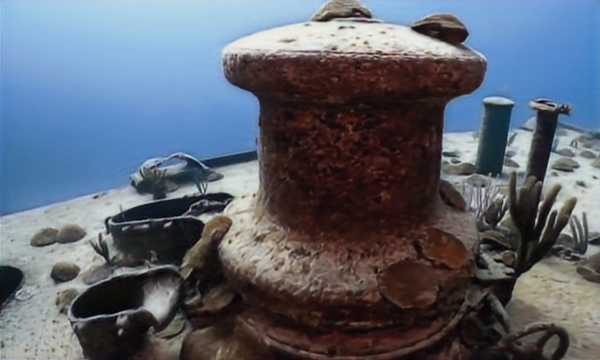}
\label{fig:debris_uies} 
\end{subfigure}%
\begin{subfigure}[t]{0.15\textwidth}
  \centering
  \includegraphics[width=1.0\linewidth,height=0.6\linewidth]{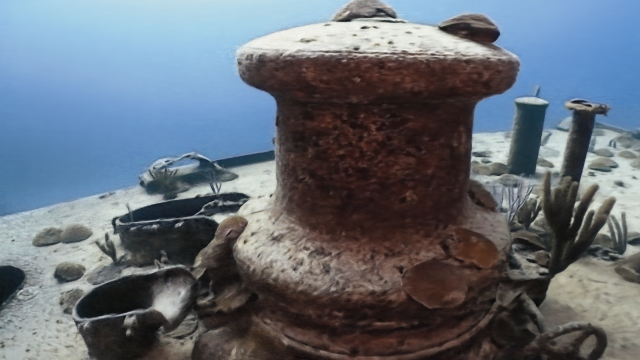}
\label{fig:debris_upifm} 
\end{subfigure}%
\begin{subfigure}[t]{0.15\textwidth}
  \centering
  \includegraphics[width=1.0\linewidth]{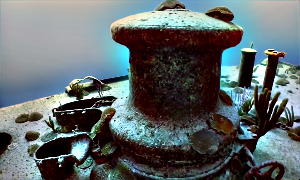}
\label{fig:debris_nerf_clean} 
\end{subfigure}%

\setcounter{subfigure}{0}
\centering

\begin{subfigure}[t]{0.15\textwidth}
  \centering
  \includegraphics[width=1.0\linewidth]{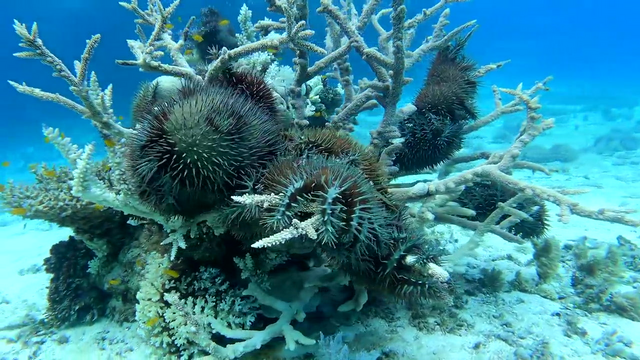}
\label{fig:debris_uw} 
\end{subfigure}%
\begin{subfigure}[t]{0.15\textwidth}
  \centering
  \includegraphics[width=1.0\linewidth]{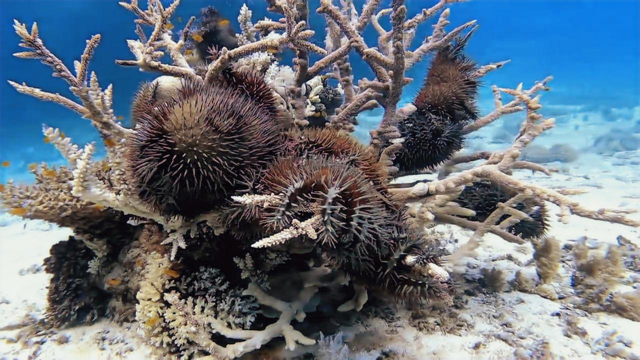}
\label{fig:debris_u2nerf} 
\end{subfigure}%
\begin{subfigure}[t]{0.158\textwidth}
  \centering
  \includegraphics[width=1.0\linewidth]{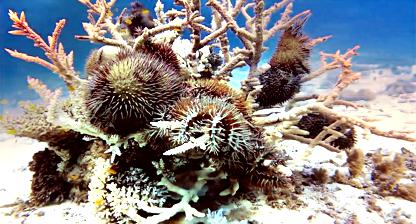}
\label{fig:debris_u2nerf} 
\end{subfigure}%
\begin{subfigure}[t]{0.15\textwidth}
  \centering
  \includegraphics[width=1.0\linewidth]{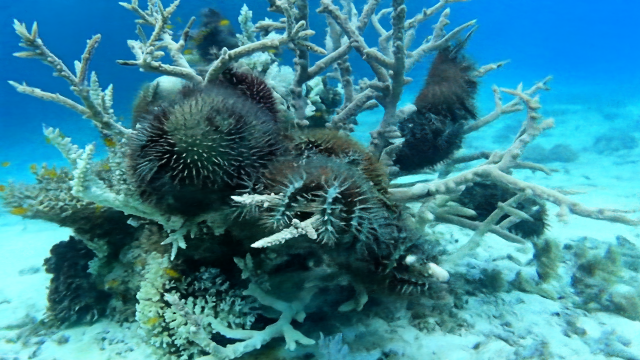}
\label{fig:debris_uw} 
\end{subfigure}%
\begin{subfigure}[t]{0.15\textwidth}
  \centering
  \includegraphics[width=1.0\linewidth]{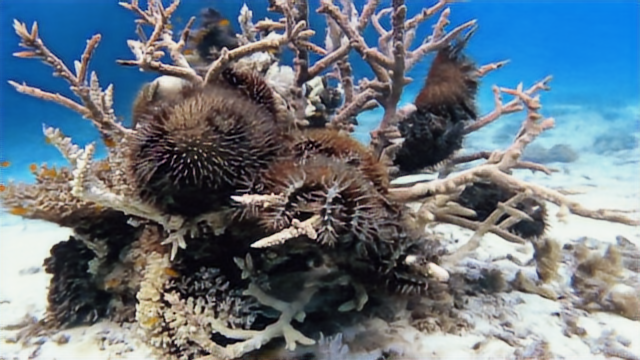}
\label{fig:debris_uies} 
\end{subfigure}%
\begin{subfigure}[t]{0.15\textwidth}
  \centering
  \includegraphics[width=1.0\linewidth]{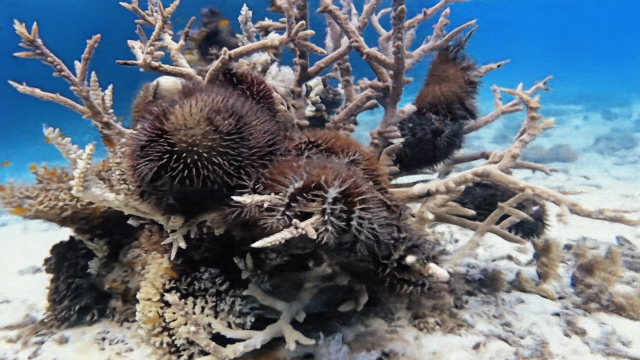}
\label{fig:debris_upifm} 
\end{subfigure}%
\begin{subfigure}[t]{0.15\textwidth}
  \centering
  \includegraphics[width=1.0\linewidth]{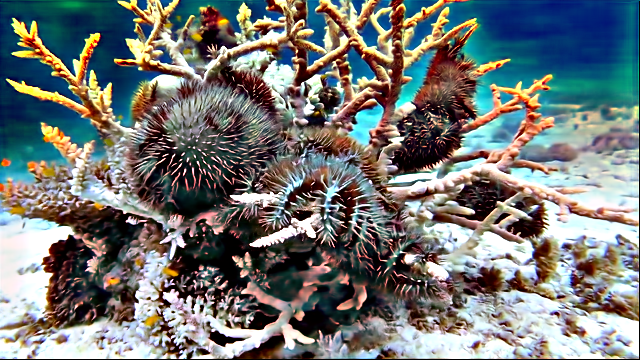}
\label{fig:debris_nerf_clean} 
\end{subfigure}%

\setcounter{subfigure}{0}
\centering

\begin{subfigure}[t]{0.15\textwidth}
  \centering
  \includegraphics[width=1.0\linewidth]{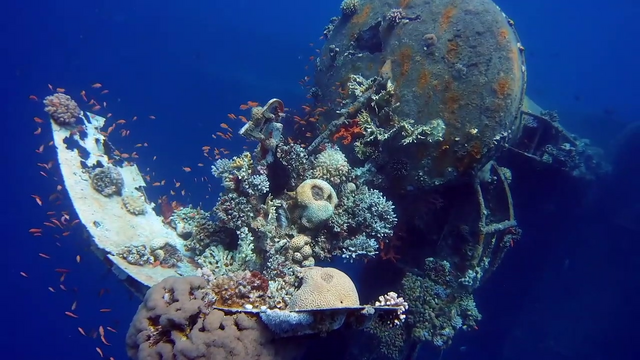}
\caption{Underwater Image}
\label{fig:debris_uw} 
\end{subfigure}%
\begin{subfigure}[t]{0.15\textwidth}
  \centering
  \includegraphics[width=1.0\linewidth]{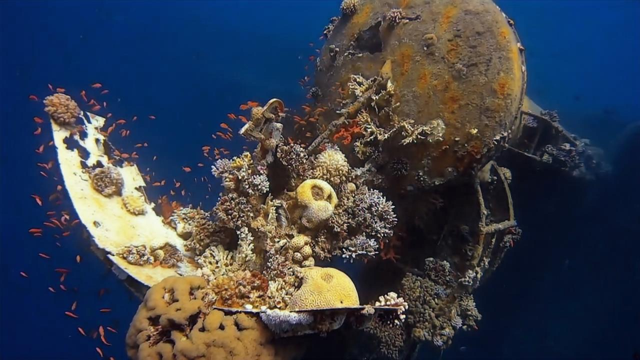}
\caption{UIESS}
\label{fig:debris_u2nerf} 
\end{subfigure}%
\begin{subfigure}[t]{0.158\textwidth}
  \centering
  \includegraphics[width=1.0\linewidth]{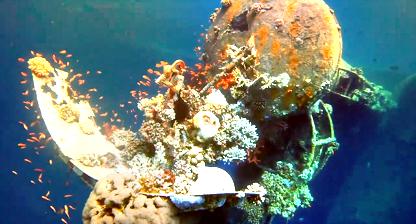}
\caption{UPIFM}
\label{fig:debris_u2nerf} 
\end{subfigure}%
\begin{subfigure}[t]{0.15\textwidth}
  \centering
  \includegraphics[width=1.0\linewidth]{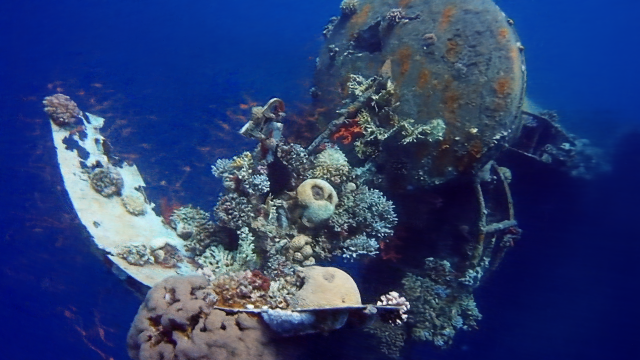}
\caption{NeRF}
\label{fig:debris_uw} 
\end{subfigure}%
\begin{subfigure}[t]{0.15\textwidth}
  \centering
  \includegraphics[width=1.0\linewidth]{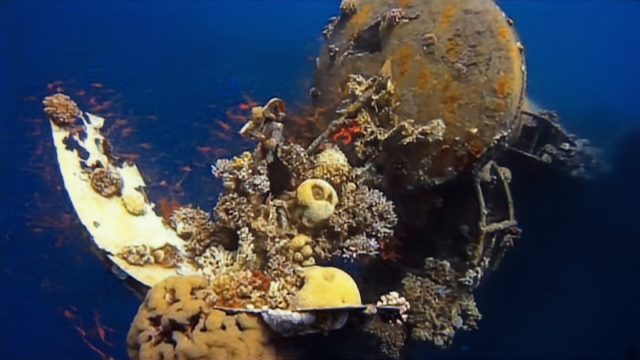}
\caption{NeRF + Clean}
\label{fig:debris_uies} 
\end{subfigure}%
\begin{subfigure}[t]{0.15\textwidth}
  \centering
  \includegraphics[width=1.0\linewidth]{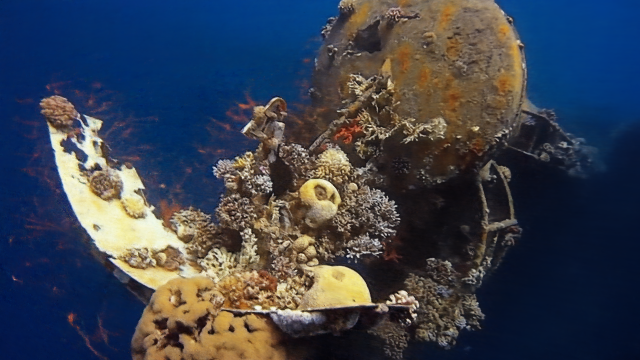}
\caption{Clean + NeRF}
\label{fig:debris_upifm} 
\end{subfigure}%
\begin{subfigure}[t]{0.15\textwidth}
  \centering
  \includegraphics[width=1.0\linewidth]{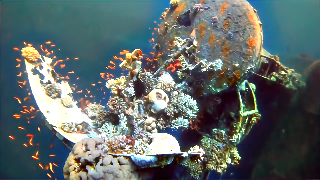}
\caption{U2NeRF}
\label{fig:debris_nerf_clean} 
\end{subfigure}%

\caption{Qualitative results on single-scene rendering for Medium and Hard Scenes. The top 4 rows depict the scenes from the hard split(Scene 1, Scene 2, Scene 3 and Scene 4 respectively) and the last 4 rows depict the scene from the medium split(coral, shipwreck, starfish and debris respectively). (a) represents the actual underwater image from the scene, (b) \& (c) represents the no rendering baseline methods (\cite{chen2022domain} \& \cite{chai2022unsupervised}), (d), (e) \& (f) refers to the renderings from NeRF on raw underwater image, restored view after NeRF rendering and NeRF rendering on restored input underwater images respectively, and (g) refers to results from our method: U2NeRF. U2NeRF is able to render better high-quality images when compared to other rendering+restoring methods.}
\label{fig:single_mh}
\end{figure*}

We provide complete Qualitative (Fig. \ref{fig:single_mh} and Fig. \ref{fig:single_e}) and Quantitative (Table. \ref{tab:hard}, Table. \ref{tab:medium}) results of our method on all 3 splits which includes 12 scenes in total. For the medium and hard splits, we display seven distinct qualitative findings, including baselines such as NeRF, Clean+NeRF, and NeRF+Clean where the restoration is carried out using \cite{chen2022domain}. We also display the qualitative outcomes of two different restoration techniques (\cite{chen2022domain} and \cite{chai2022unsupervised}). For the easy split, we display baselines such as NeRF and NeRF+Clean as well as outcomes from two restoration techniques. We use \cite{desai2022aquagan} to degrade the LLFF\cite{mildenhall2019llff} data to create the easy split.  

\section{Limitations/Future Work}
In the hard split, we particularly observe the blurriness in the regions of the rendered image where there is a significant movement of plants. This opens up some intriguing possibilities for the future, particularly in terms of solving the issue of object motion.

\clearpage

\begin{table*}
  \resizebox{0.99\textwidth}{!}{
  \begin{tabular}{llccc|ccc|ccc|ccc}
    \toprule
    \multirow{2}{*}{Setting} & \multirow{2}{*}{Models} & \multicolumn{3}{c|}{Scene 1}  & \multicolumn{3}{c|}{Scene 2} & \multicolumn{3}{c|}{Scene 3} & \multicolumn{3}{c}{Scene 4}\\
    \cmidrule(r){3-14}
    & & UIQM$\uparrow$ & UCIQE$\uparrow$ & LPIPS$\downarrow$ 
    & UIQM$\uparrow$ & UCIQE$\uparrow$ & LPIPS$\downarrow$ & UIQM$\uparrow$ & UCIQE$\uparrow$ & LPIPS$\downarrow$ & UIQM$\uparrow$ & UCIQE$\uparrow$ & LPIPS$\downarrow$\\
    \midrule
    \multirow{2}{*}{No Rendering} & UIESS & 0.63 & 25.19 & - & 0.68 & 28.06 & - &  0.57 & 29.85 & - & 0.70 & 25.53 & -\\
    & UPIFM & 0.89 & 23.03 & - & 1.19 & 29.80 & - & 1.23 & 30.75 & - & 1.41 & 30.56 & -\\
    \midrule
    \multirow{4}{*}{Rendering} & NeRF & 0.32 & 14.01 & 0.26 & 0.68 & 22.91 & 0.32 & 0.32 & 20.48 & 0.31 & 0.51 & 16.06 & 0.43\\
    & NeRF+Clean & 0.50 & \textbf{25.82} & 0.25 & 0.53 & 27.84 & 0.32 & 0.34 & \textbf{30.06} & 0.30 & 0.57 & \textbf{26.08} & 0.42\\
    & Clean+NeRF & 0.44 & 24.37 & \textbf{0.21} & 0.49 & 27.62 & 0.30 & 0.35 & 29.42 & 0.30 & 0.52 & 24.69 & \textbf{0.34}\\
    \cmidrule(r){2-14}
    & U2NeRF & \textbf{0.84} & 23.33 & 0.22 & \textbf{1.32} & \textbf{30.42} & \textbf{0.21} & \textbf{1.04} & 29.60 & \textbf{0.22} & \textbf{1.18} & 23.80 & 0.37\\
    \bottomrule
  \end{tabular}}
  \caption{Comparison of U2NeRF against SOTA methods for single scene rendering on the UVS Dataset, Hard Split (scene-wise).}
  \label{tab:hard}
\end{table*}

\begin{table*}
  \resizebox{0.99\textwidth}{!}{
  \begin{tabular}{llccc|ccc|ccc|ccc}
    \toprule
    \multirow{2}{*}{Setting} & \multirow{2}{*}{Models} & \multicolumn{3}{c|}{Coral}  & \multicolumn{3}{c|}{Debris} & \multicolumn{3}{c|}{Starfish} & \multicolumn{3}{c}{Shipwreck}\\
    \cmidrule(r){3-14}
    & & UIQM$\uparrow$ & UCIQE$\uparrow$ & LPIPS$\downarrow$ 
    & UIQM$\uparrow$ & UCIQE$\uparrow$ & LPIPS$\downarrow$ & UIQM$\uparrow$ & UCIQE$\uparrow$ & LPIPS$\downarrow$ & UIQM$\uparrow$ & UCIQE$\uparrow$ & LPIPS$\downarrow$\\
    \midrule
    \multirow{2}{*}{No Rendering} & UIESS & 1.11 & 28.11 & - & 0.84 & 31.83 & - & 1.61 & 32.55 & - & 0.98 & 29.63 & -\\
    & UPIFM & 1.30 & 32.05 & - & 1.11 & 32.09 & - & 1.91 & 33.85 & - & 1.35 & 33.75 & -\\
    \midrule
    \multirow{4}{*}{Rendering} & NeRF & 0.19 & 28.19 & 0.22 & 0.50 & \textbf{34.45} & 0.175 & 0.94 & 31.60 & 0.21 & 0.35 & \textbf{32.23} & 0.21\\
    & NeRF+Clean & 0.71 & 28.01 & 0.23 & 0.70 & 32.57 & 0.20 & 1.34 & 32.96 & 0.23 & 0.70 & 30.65 & 0.22\\
    & Clean+NeRF & 0.69 & 27.93 & 0.21 & 0.68 & 31.51 & 0.175 & 1.33 & 32.50 & 0.20 & 0.71 & 29.39 & 0.20\\
    \cmidrule(r){2-14}
    & U2NeRF & \textbf{1.34} & \textbf{31.17} & \textbf{0.16} & \textbf{1.17} & 32.08 & \textbf{0.17} & \textbf{2.22} & \textbf{35.12} & \textbf{0.18} & \textbf{1.54} & 31.83 & \textbf{0.17}\\
    \bottomrule
  \end{tabular}}
  \caption{Comparison of U2NeRF against SOTA methods for single scene rendering on the UVS Dataset, Medium Split (scene-wise).}
  \label{tab:medium}
\end{table*}

\end{document}